\documentclass[10pt,twocolumn,letterpaper]{article}

\usepackage[pagenumbers]{cvpr} 

\usepackage{algorithm}
\usepackage{algpseudocode}
\usepackage{fancyvrb}
\usepackage{fvextra}
\usepackage{makecell}
\usepackage{multirow}
\usepackage{pifont}
\usepackage[most]{tcolorbox}
\usepackage{wrapfig}
\usepackage[table]{xcolor}

\definecolor{catcolor}{RGB}{240,245,255}

\newtcolorbox[auto counter]{promptbox}[2][]{
  enhanced,
  breakable,
  colback=gray!5,
  colframe=gray!40,
  coltitle=black,
  title={Prompt~\thetcbcounter: #2},
  #1 
}

\definecolor{cvprblue}{rgb}{0.21,0.49,0.74}
\usepackage[pagebackref,breaklinks,colorlinks,allcolors=cvprblue]{hyperref}

\def\paperID{*****}
\def\confName{3DV\xspace}
\def\confYear{2027\xspace}

\title{Task-Adaptive Grounded 3D-Programmers Using 2D VLMs}

\author{Arman Raayatsanati\textsuperscript{1,\textasteriskcentered,\textdagger} \quad Sombit Dey\textsuperscript{2,\textasteriskcentered} \quad Anna-Maria Halacheva\textsuperscript{2,\textasteriskcentered} \quad Jan-Nico Zaech\textsuperscript{2} \\
Luc Van Gool\textsuperscript{2} \quad Danda Pani Paudel\textsuperscript{2} \\
\textsuperscript{1}TextQL \quad \textsuperscript{2}INSAIT, Sofia University ``St.\ Kliment Ohridski''
}

\begin{document}

\twocolumn[{%
\renewcommand\twocolumn[1][]{#1}%
\maketitle
\begin{center}
    \captionsetup{type=figure}
    \includegraphics[width=\textwidth,trim={0cm 0.3cm 0cm 0cm}]{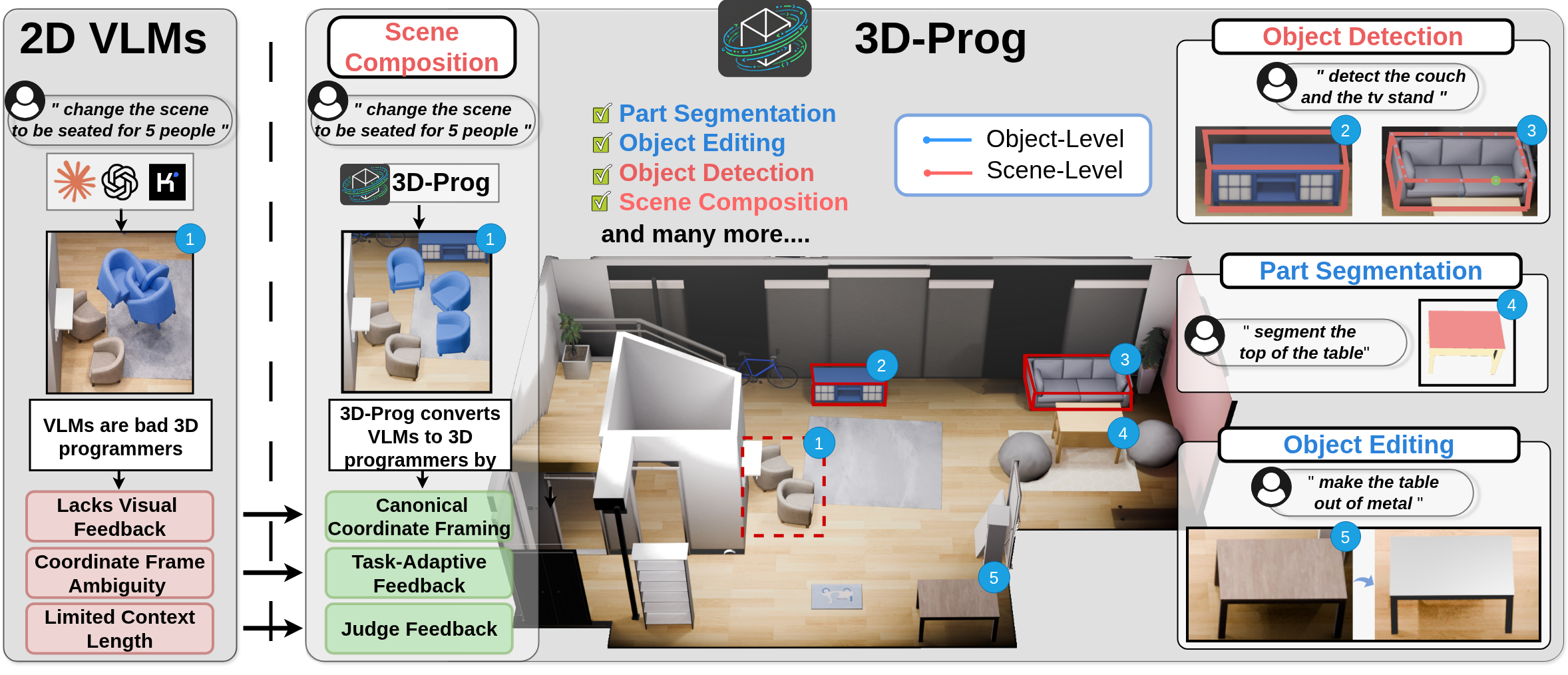}
    \captionof{figure}{%
    Today, 2D VLMs lack the zero-shot capacity to generate, modify, or reason about 3D assets.
    To bridge this gap, we introduce \textbf{3D-Prog}, a modular and flexible framework for 3D visual programming.
    By leveraging the semantic reasoning of 2D VLMs, 3D-Prog acts as a \emph{3D programmer} that interprets multi-view 2D visualizations of a 3D asset within a canonical frame.
    The \textbf{3D-Prog} approach addresses a broad range of 3D tasks spanning generation, understanding, and compositional reasoning.%
    }
    \label{fig:teaser_fig}
\end{center}%
}]

\makeatletter
\iftoggle{cvprfinal}{%
  {\renewcommand{\thefootnote}{\fnsymbol{footnote}}%
   \renewcommand{\@makefntext}[1]{\noindent\@makefnmark#1}%
   \footnotetext[1]{Equal contribution.}%
   \footnotetext[2]{Research done as part of an ETH Zurich MSc thesis hosted by INSAIT.}}%
}{}
\makeatother

\begin{abstract}
Recent vision-language models (VLMs) exhibit remarkable generalization and reasoning abilities, yet 3D understanding in these models is limited by data scale, training diversity, and reasoning capacity. Instead of naively extending these models into 3D, we take a different approach: we enable powerful 2D VLMs to operate reliably in 3D by introducing 3D grounding and iterative feedback loops with two novel concepts: Canonical Coordinate Framing (CCF) and Task-Adaptive Feedback (TAF).
CCF serves as a unified visual representation that anchors both inputs and outputs to a shared Euclidean coordinate system, solving common challenges in 3D grounding such as axis ambiguity, inconsistent metric scale, and floating references. Complementary to this structured framing of the 3D inputs, TAF closes the reasoning loop with task-adaptive dynamic feedback that enables 2D VLMs to perform varied open-vocabulary tasks within their native visual context.

Building on this foundation, we introduce 3D-Prog, a 3D understanding, reasoning, and generation framework that jointly employs the capabilities of CCF and TAF together with powerful VLMs. Without requiring any retraining, 3D-Prog performs open-vocabulary 3D understanding, manipulation, and generation across both object-level and scene-level tasks. Our experiments show that the joint use of CCF and TAF transforms 2D VLMs into geometry-aware 3D programmers, achieving consistent, interpretable, and high-quality results across diverse 3D tasks.
\end{abstract}
   
\section{Introduction}
The pursuit of general-purpose 3D understanding and manipulation systems has traditionally relied on specialized models, each tailored for a specific geometric task~\cite{Kim2024PartSTAD2P, Zhao2024SweepNetUL, Liu2024Part123P3}. Whether converting scanned objects into simulation-ready models, segmenting parts, or composing scenes from language, these systems depend on extensive 3D annotations, domain expertise, and task-specific training \cite{Peng2023OpenScene,jia2024sceneverse,Halacheva2024HolisticUO,shao2025great}. This specialization fragments the 3D ecosystem and limits scalability to diverse real-world 3D tasks in robotics, graphics, and vision.

Meanwhile, vision-language models (VLMs) trained on massive 2D image-text corpora have demonstrated remarkable generalization, compositional reasoning, and zero-shot learning abilities \cite{tschannen2025siglip,deitke2025molmo,comanici2025gemini}. Yet, despite their semantic power, these models operate within the 2D image manifold and lack explicit grounding in metric 3D space \cite{yang2025thinking}. Emerging 3D VLMs attempt to bridge this gap, but their generalization remains constrained by limited data scale and diversity \cite{zhi2025lscenellm,khanna2024habitat}. As a result, neither approach provides sufficient performance; directly prompting 2D VLMs for 3D quantities leads to unstable and inconsistent predictions \cite{yang2025thinking}, while 3D VLMs lack a diversity of capabilities~\cite{zhi2025lscenellm}. 

We address this challenge by introducing a task-adaptive and grounded 3D programming system built on two novel VLM communication mechanisms: \textbf{Canonical Coordinate Framing} (CCF) for 3D grounding and \textbf{Task-Adaptive Feedback} (TAF).
CCF anchors both inputs and outputs to a shared Euclidean coordinate system by rendering multi-view orthographic projections with explicit, visible axes, metric scale marks, and a fixed global origin. These visual cues eliminate what we call coordinate-frame ambiguity and allow 2D VLMs to perform spatial reasoning within a consistent geometric frame.
While spatial grounding is necessary, it is often insufficient for complex tasks. To address this, we propose TAF, which provides an iterative execution-based feedback loop that serves as a versatile reasoning scratch-space for task-specific information, from physical material descriptions to layout reasoning.
Unlike approaches that attempt to merely extend VLMs into 3D with limited data, the combination of CCF and TAF empowers existing 2D models to reason metrically by grounding 3D understanding visually, without retraining or architectural modification. This design transforms pretrained 2D VLMs into geometry-aware systems capable of interpreting, editing, and generating 3D content.

Building on this foundation, we introduce \textbf{3D-Prog}, a 3D reasoning and generation framework that utilizes our task-adaptive and grounded 3D representation. 3D-Prog takes 3D assets such as point clouds, meshes, or Universal Scene Description (USD) scenes along with a natural-language instruction, carries out task-specific analysis, and generates a structured 3D manipulation program. Each line in the program invokes modules for geometric analysis, 3D manipulation, or visual feedback in canonical coordinates. The generated programs consume and produce intermediate results, forming self-contained reasoning chains that are interpretable and executable. 
The outputs are written in the USD format, providing a modular, hierarchical, and interpretable representation that is naturally compatible with both 2D VLMs and 3D editing workflows.

Without any retraining or fine-tuning, 3D-Prog performs remarkably well on {open-vocabulary 3D understanding, manipulation, and synthesis across object-level and scene-level tasks}. It unifies 3D perception, reasoning, and generation in a visually grounded framework, demonstrating that \textbf{explicit canonical framing is a prerequisite} for robust geometry-aware multimodal reasoning.

Our key contributions are summarized as follows:
\begin{itemize}
\item \emph{Canonical Coordinate Framing (CCF)}: a representation that visually grounds 3D reasoning for 2D VLMs by enforcing a shared Euclidean coordinate system for inputs and outputs, resolving the coordinate-frame ambiguity.

\item \emph{Task-Adaptive Feedback (TAF)}: a reasoning scratch-space that enables a CCF-guided VLM to break out of purely geometric reasoning and perform generalist, task-adaptive reasoning over 3D assets. 

\item \emph{3D-Prog}: a 3D understanding, reasoning, and generation framework building on CCF and TAF. We show that 2D VLMs embedded in 3D-Prog achieve consistent, interpretable, and high-quality results across diverse 3D tasks, without task-specific training or fine-tuning. 
\end{itemize}

\begin{figure*}
\centering
\includegraphics[width=0.8\linewidth]{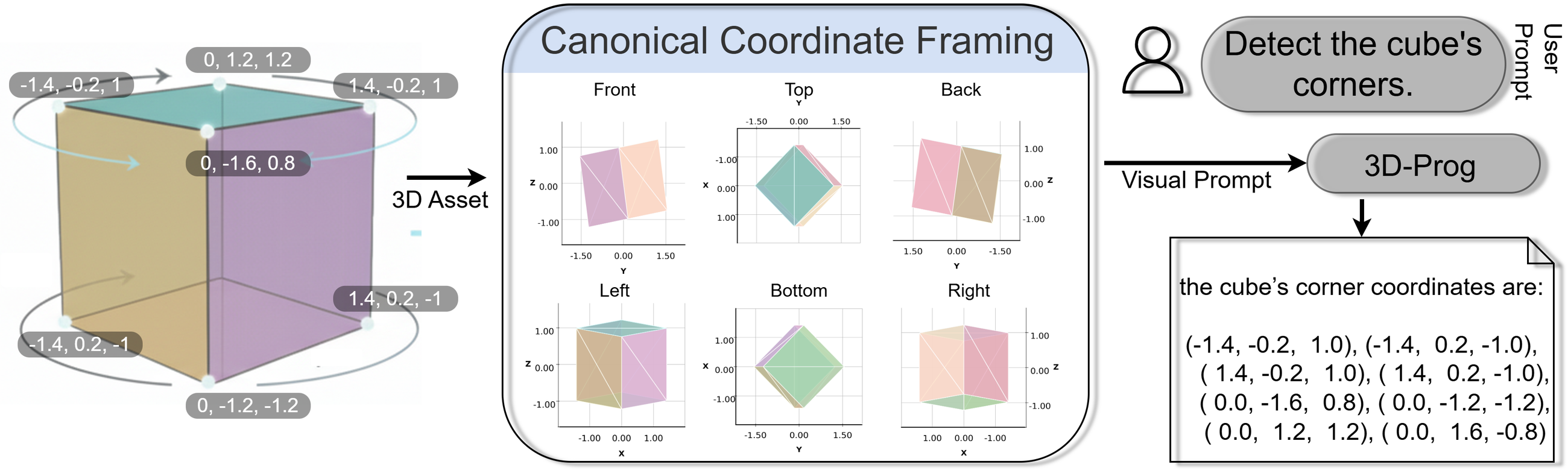}
\caption{
\label{fig:toyexample}
\textbf{CCF Overview.} The CCF transforms an input 3D asset into a spatially grounded 2D visual prompt by rendering its six orthographic views, each featuring a fixed origin, visible axes, and metric scale marks.}
\end{figure*}
\section{Related Work}
\label{sec:relatedwork}
\textbf{2D VLMs and Emergent Spatial Understanding.}
Vision-language models (VLMs) \cite{deitke2025molmo,tschannen2025siglip,hurst2024gp4,comanici2025gemini,anthropic2025introducingClaude4} have demonstrated remarkable multimodal reasoning capabilities across a wide range of 2D tasks, from image captioning to open-world recognition and VQA. 
Despite being trained on 2D image-text pairs, these models exhibit emergent spatial and physical understanding: they can ground language in object relations, infer affordances, and even support world models and embodied reasoning frameworks \cite{bruce2024genie,black2024pi_0,octo_2023,kim2024openvla,ma2022sqa3d}. 
However, their spatial reasoning remains implicitly encoded within 2D coordinates. 
As recent evaluations highlight~\cite{Chen_2024_CVPR,yang2025thinking,ma2022sqa3d, huang2023embodied}, these models lack explicit 3D consistency and cannot directly produce geometry-grounded outputs, limiting their use in 3D reasoning and manipulation.

\textbf{3D-aware VLMs: Bridging to 3D Understanding.}
To overcome these limitations, several works have introduced 3D-aware VLMs that explicitly model spatial structure or leverage 3D supervision. 
Some approaches employ 3D backbones \cite{gaussianvlm,mei2025perla,chen2024ll3da,huang2024chat,zhu20233d,3dllm}, e.g., point transformers, while others extend 2D models, e.g., with multi-view consistency \cite{li2025triposg,hunyuan3d22025tencent,Ha2022SemanticAO,zhang2022pointclip,conceptfusion}.
These methods achieve strong grounding in 3D perception and have been applied to both understanding tasks (e.g., 3D captioning, grounding, QA) and generation tasks (e.g., text-to-3D synthesis, scene reconstruction).
Despite their promising progress, these 3D-aware VLMs remain constrained by limited multimodal 3D datasets and task-specific training objectives, which hinder their ability to generalize across domains and tasks as broadly as 2D VLMs do \cite{wang2025vigil3d}.
Moreover, existing 3D VLMs typically specialize in either language understanding of 3D assets (e.g., captioning, grounding, VQA) \cite{gaussianvlm,mei2025perla,chen2024ll3da,huang2024chat,zhu2024unifying,li2025m3dbench} or 3D output generation (e.g., text-to-3D synthesis) \cite{li2025triposg,hunyuan3d22025tencent}, rather than offering a unified and flexible framework for both understanding and manipulation.

\textbf{Visual Programming and Programmatic Reasoning.}
Beyond model-centric approaches, visual programming frameworks have emerged as a complementary paradigm for integrating perception, reasoning, and action through modular composition.
Neural Module Networks~\cite{Andreas2015NeuralMN} pioneered this direction by decomposing visual question answering into structured, executable programs.
Building on this idea, recent works such as VisProg \cite{Gupta2022VisualPC}, ViperGPT \cite{Suris2023ViperGPTVI}, and VADAR \cite{Marsili_2025_CVPR} demonstrate that complex visual tasks can be represented as modular programs, differing mainly in their approach to program synthesis and API generation.
However, this entire line of work remains focused on 2D vision tasks (e.g., detection, VQA, image editing), and thus cannot ingest or operate directly on 3D visual inputs.
Efforts towards 3D visual program synthesis have so far been domain-specific, targeting single tasks \cite{yuan2024cadtalk,littlefair2025flairgpt,SpatialLM} or specialized 3D asset types \cite{mallis2025cad}, which limits their generality. 
In contrast, 3D-Prog~supports diverse 3D assets spanning synthetic and scanned data across varying levels of granularity (object- and scene-level),
and enables a wide range of 3D tasks, including generation, editing, and understanding.

\section{Method}
This section presents \textit{3D-Prog}, our approach for converting 2D VLMs into 3D programmers capable of generation, understanding, manipulation, and reasoning over 3D assets. To enable 2D VLMs to interact with 3D assets, we first propose a 3D framing protocol, \textbf{CCF}, aimed at grounding the 3D inputs to a canonical Euclidean space via images, as explained in~\cref{subsec:3dgrounding}.
To support an open-ended range of 3D tasks, we further propose \textbf{TAF}, which allows the model to request and utilize dynamic visual feedback tailored to the instruction, explained in~\cref{subsec:TAF}.
Finally,~\cref{subsec:3Dprog} presents the \textbf{3D-Prog} framework, which integrates these components into a unified system capable of performing diverse and complex 3D programming tasks.

\begin{figure*}[t]
    \centering
    \includegraphics[width=\linewidth]{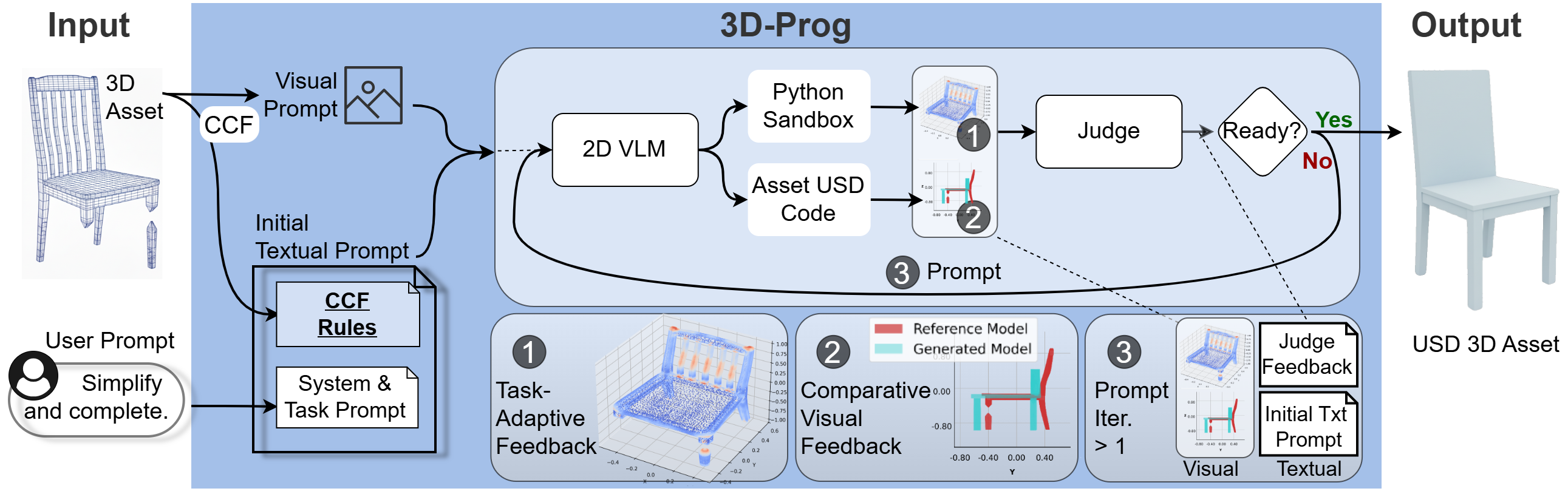}
    \caption{\textbf{3D-Prog Overview.} Before being passed to 3D-Prog, the user input is pre-processed by the CCF. Within 3D-Prog, a 2D VLM generates Python code for both (1) Task-Adaptive Analysis and (2) asset generation. A Judge module evaluates this feedback and either confirms the task is complete or provides (3) Iterative Prompt Feedback, comprising both visual and textual cues, which is fed back to the 2D VLM for iterative refinement.}
    \label{fig:method}
\end{figure*}

\subsection{Canonical Coordinate Framing}
\label{subsec:3dgrounding}

\textbf{CCF} is a 3D grounding representation strategy that visually anchors both 3D inputs and outputs within a shared global Euclidean coordinate system, as illustrated in~\cref{fig:toyexample}.
Specifically, we render a single visualization with six orthographic projections of the 3D asset, along with a supplementary perspective view for spatial context. The projections share \textbf{a)} visible and labeled $X$, $Y$, and $Z$ axes, \textbf{b)} metric scale cues via axis markings, and \textbf{c)} a fixed reference origin anchored to the asset's bounding box across all views.

By representing 3D geometry through consistent multi-view visual-linguistic cues, CCF removes gauge ambiguities and establishes a stable and grounded spatial reference for reasoning. 
CCF shares its principles with various geometric methods such as DUSt3R~\cite{Wang_2024_CVPR}, MASt3R~\cite{mast3r_eccv24}, and VGGT~\cite{Wang_2025_CVPR} to achieve multi-view consistency in a shared canonical space, enforcing coherence across viewpoints. 
Within this canonical frame, the VLM is prompted to express predicted 3D coordinates and spatial relationships directly.

\begin{table}[t]
\centering
\small
\setlength{\tabcolsep}{4pt}
\caption{\textbf{Failure analysis of ungrounded designs.} Explicit metric grounding is essential for stable and interpretable 3D reasoning.}
\begin{tabular}{l l}
\toprule
\textbf{Missing Design Element} & \textbf{Observed Failure Mode} \\
\midrule
No visible axes & Orientation ambiguity \\
No axis markings or values & Inconsistent scale \\
Perspective-only projections & Metric distortion \\
Parametric-only 3D outputs & Floating reference frame \\
\bottomrule
\end{tabular}
\label{tab:failures}
\end{table}

\textit{Why Do Other Designs Fail?}
We observe that designs lacking explicit metric grounding exhibit degenerate behaviors, as summarized in~\cref{tab:failures}.
Without a visible coordinate frame, VLMs struggle to maintain orientation, scale, and metric consistency across views.

By introducing a visible and shared Euclidean reference, \textbf{CCF} transforms pretrained 2D VLMs into reliable 3D reasoning engines.
It enables a) consistent spatial queries (distance, adjacency, containment), b) geometry-aware visual grounding across dialogue turns, c) controllable and interpretable 3D scene editing, and d) multi-view alignment without pose metadata.

We conclude that {3D grounding through explicit canonical framing is a prerequisite for robust geometry-aware multimodal interaction}.

\subsection{Task-Adaptive Feedback}
\label{subsec:TAF}

A key capability of a generalist 3D programmer is the ability to request and obtain task-relevant feedback.
While CCF enables seamless interaction between 3D assets and VLMs, the generated asset may still fall short of the task requirements.

To bridge this gap, we propose \textbf{TAF}, a mechanism that establishes a task-dependent feedback loop between the VLM and its generated content. Concretely, we link the VLM to a Python execution environment, enabling it to produce TAF as visual feedback. In each iteration, the VLM generates USD manipulation code and task-specific analysis code that programmatically inspects the current 3D asset, producing diagnostic visualizations (e.g., height distributions, density heatmaps, segmentation overlays).
These visual outputs are rendered and fed back to the VLM alongside the updated CCF views of the generated asset, forming an iterative loop in which the VLM identifies discrepancies, refines its hypotheses, and corrects its output based on concrete evidence from the asset itself. The VLM autonomously determines what intermediate analysis is needed for a given task, allowing TAF to range from plotting material properties and assessing physical plausibility to running non-trivial operations such as DBSCAN clustering to isolate object parts for segmentation (see~\cref{fig:task_adaptive}).

\subsection{3D-Prog Framework}
\label{subsec:3Dprog}

Building upon CCF and TAF, we introduce 3D-Prog, a generalist 3D programming framework that leverages the broad semantic and reasoning capabilities of the 2D VLM to operate effectively on 3D assets.
3D-Prog transforms VLMs into 3D programmers capable of carrying out a wide range of tasks without requiring task-specific supervision or adaptation.
Our framework is designed around three core principles: (1) CCF expresses 3D assets in a canonical frame, providing grounding cues that enable VLMs trained on 2D data to infer spatial relationships; (2) TAF equips the system with an iterative feedback mechanism that closes the loop between generation and evaluation; and (3) the USD format, with its hierarchical and compositional nature, aligns scene representation with VLM reasoning patterns, enabling complex 3D transformations to be decomposed into interpretable programmatic steps.

Following these principles, the architecture comprises three primary modules:

\textbf{2D VLM for perception}:
CCF-rendered views of the 3D assets serve as the perceptual input to the VLM. Conditioned on these renders and the task description, the VLM produces a) semantic and geometric interpretations of the scene and
b) executable code that governs the manipulation of the 3D asset. The generated code is executed inside a Python environment, which provides the essential capabilities of TAF and USD manipulation, enabling the VLM to construct, edit, or transform the underlying USD scene via OpenUSD in Python.

\textbf{Judge}: To provide critical feedback, a separate {Judge VLM} evaluates the proposed solution. The generated USD asset is rendered and overlaid with the reference model’s CCF representation (\cref{subsec:3dgrounding}) to provide iterative visual feedback to the VLM. The judge evaluates the solution based on the user task prompt, comparative visual feedback, and TAF-produced visual analysis. The evaluation yields a structured response comprising a binary completion status, a confidence score, and a diagnostic report (a feedback comment alongside lists of detected errors and prescriptive suggestions). The judge response is appended to the Python output to guide the VLM in subsequent iterations.

\textbf{Iterative Refinement through TAF}: The judge feedback and TAF-produced diagnostic visualizations are fed back to the VLM, which generates an updated USD asset in the next iteration. This TAF-driven loop allows the VLM to correct trivial mistakes (such as faulty USD code, incorrect axes, or wrong rotations), improve geometric fidelity based on visual evidence from its own analysis, and progressively refine its output from coarse to fine-grained solutions. The process continues until the task is successfully completed or a predefined maximum number of iterations is reached.

\subsection{Generalization of 3D-Prog}
\label{sec:tasks}
3D-Prog handles a wide range of 3D vision tasks within a unified framework, spanning three functional domains: \textbf{generation} (object completion, Scan2Sim), \textbf{understanding} (detection, segmentation, editing, QA, captioning), and \textbf{reasoning} (compositional scene building, articulation prediction). A detailed task breakdown is provided in the Supplementary Material, Tab.~\ref{tab:tasks}.

By expressing all outputs as USD code, diverse 3D objectives become instances of the same composable representation that can be inspected, edited, and executed across standard USD-based toolchains. 
This unified abstraction allows a single system to analyze, modify, and construct 3D scenes from natural-language or visual instructions, positioning USD as a lingua franca for 3D understanding and generation. 
The range of tasks spans both synthetic and real-world 3D assets at object and scene scales, demonstrating the generalization of 3D-Prog across asset types.

Within 3D generation, the system performs tasks that synthesize or transform raw geometry into simulation-ready assets, such as object completion and ``Scan2Sim'' reconstruction. To facilitate 3D understanding, 3D-Prog interprets and structures data across geometrical, spatial, visual, contextual, and semantic-spatial dimensions, enabling tasks such as object detection or object editing. Finally, reasoning tasks are a core capability, e.g., predicting articulation and composing coherent scenes from discrete objects.
\section{Experiments}
\label{sec:experiments}

\begin{table}[t]
\centering
\scriptsize
\setlength{\tabcolsep}{3pt}
\renewcommand{\arraystretch}{1.15}
\caption{Comparison of 3D-Prog with closed (Gemini 2.5 Pro, GPT-5.2, Claude Sonnet 4.5) and open (Kimi K2.5) 2D VLMs on generation, understanding, and reasoning tasks in 3D. 3D-Prog (C) and (K) use Claude and Kimi as backbones. \ding{55}~denotes failure to produce valid USD outputs.}
\resizebox{\linewidth}{!}{%
\begin{tabular}{l cccc cc}
\toprule
& \multicolumn{4}{c}{\textbf{Closed Models}} & \multicolumn{2}{c}{\textbf{Open Models}} \\
\cmidrule(lr){2-5} \cmidrule(lr){6-7}
\textbf{Task}
& Gemini & GPT & Claude & \makecell{3D-Prog\\(C)}
& Kimi & \makecell{3D-Prog\\(K)} \\
\midrule
\rowcolor{catcolor}
\multicolumn{7}{c}{Generation} \\
Object Compl. (Chamfer $\downarrow$)
& 1.17 & 0.51 & 0.42 & \textbf{0.19} & 0.45 & \textbf{0.16} \\
Scan2Sim (Chamfer $\downarrow$)
& \ding{55} & \ding{55} & \ding{55} & \textbf{0.08} & \ding{55} & \textbf{0.09} \\
\rowcolor{catcolor}
\multicolumn{7}{c}{Understanding} \\
Object Det. (mAP $\uparrow$)
& 0.00 & 0.16 & 0.04 & \textbf{0.38} & 0.02 & \textbf{0.04} \\
Instance Seg. (mAP $\uparrow$)
& 0.18 & 0.28 & 0.21 & \textbf{0.42} & 0.22 & \textbf{0.43} \\
Semantic Seg. (mIoU $\uparrow$)
& 0.48 & 0.47 & 0.48 & \textbf{0.53} & 0.47 & \textbf{0.52} \\
Object Editing (User $\uparrow$)
& 1.40 & 1.40 & 4.00 & \textbf{4.80} & \textbf{3.59} & \textbf{3.59} \\
\rowcolor{catcolor}
\multicolumn{7}{c}{Reasoning} \\
Composition (User $\uparrow$)
& 2.70 & 2.35 & 2.00 & \textbf{4.60} & 3.00 & \textbf{3.70} \\
Articulation (Axis Err $\downarrow$)
& 86.58 & 90.00 & 86.94 & \textbf{3.50} & 89.61 & \textbf{0.64} \\
\bottomrule
\end{tabular}%
}
\label{tab:task_results}
\end{table}
We evaluate 3D-Prog through task benchmarks, ablation studies, experiments across diverse VLM families, and spatial robustness analysis.
We demonstrate that 2D VLMs, when grounded via CCF and paired with TAF, can perform diverse 3D programming tasks effectively.

\textbf{Datasets.} We evaluate 3D-Prog on diverse 3D programming tasks spanning generation, understanding, and reasoning (\cref{tab:tasks}), covering both object- and scene-level settings with synthetic and real-world data.
For Scan2Sim and articulation prediction, we use real-world scans and annotations from the Articulate3D dataset~\cite{Halacheva2024HolisticUO} at the object and scene levels, respectively.
Semantic and instance part segmentation, object editing, and object completion use synthetic assets from Objaverse-XL~\cite{deitke2023objaverse} and Pixar's OpenUSD asset library~\cite{openusd_kitchen_set}, which provide ground-truth part segmentations and semantic labels.
For object editing, we curate assets to maximize geometric and visual complexity; for object completion, inputs are generated by removing pre-segmented object parts.
For object detection, we combine synthetic data~\cite{openusd_kitchen_set} with high-quality laser scans from ScanNet++~\cite{yeshwanthliu2023scannetppSup}, which includes human-annotated segmentations.
Finally, compositional reasoning is evaluated using five natural-language queries describing complex multi-object indoor environments. The full system and judge prompts are provided in the supplementary material.

\textbf{Metrics.}
We evaluate 3D-Prog using task-specific metrics as well as human judgment to capture both geometric fidelity and semantic plausibility. 
For object-level perception, we report mean Average Precision (mAP) for detection and instance segmentation and mean Intersection over Union (mIoU) for semantic segmentation. Geometric reconstruction quality for object completion and Scan2Sim is quantified using Chamfer distance, while articulation accuracy is measured via the median motion axis error.
For object editing and compositional reasoning, where automated metrics often fail to capture nuance, we conduct a user study with four participants. 
Each instance is evaluated against five task-specific success criteria and scored on a $0$–$5$ scale. We report the mean score aggregated across all samples and evaluators. Detailed success criteria and user study protocols are provided in the supplementary material (\cref{sec:user_study}).

\textbf{Setup.}
We evaluate task performance across a diverse selection of 2D VLMs, including the proprietary systems Claude Sonnet 4.5~\cite{anthropic2025introducingClaude4}, Gemini 2.5 Pro~\cite{comanici2025gemini}, and GPT-5.2~\cite{openai2025gpt52}, as well as the open-source Kimi K2.5~\cite{kimi-k2-5} model. Additionally, we instantiate open-source and closed-source variants of 3D-Prog using the top-performing open-source and proprietary VLMs as their respective foundational cores. The iterative refinement process is capped at $20$ iterations, with an early stopping condition triggered upon approval by the judge. If the iteration limit is reached, the terminal output is recorded as the final result. Performance for each task is quantified as the mean score across five distinct assets.

\textbf{Baselines and Ablations.} In the main comparison (\cref{tab:task_results}), we evaluate four standalone 2D VLMs as baselines: Gemini 2.5 Pro, GPT-5.2, Claude Sonnet 4.5, and Kimi K2.5. We instantiate 3D-Prog with Claude and Kimi as backbones, as they are the overall top-performing closed-source and open-source baselines, respectively. The primary distinction between a baseline and its 3D-Prog variant is the omission of CCF and TAF; baselines are otherwise granted identical task prompts. Claude Sonnet 4.5 is additionally used as the dedicated baseline for ablation studies, robustness testing, and 3D VLM comparisons, as it demonstrated the highest aggregate performance among all evaluated 2D VLMs.
 
\begin{figure*}[tb]
  \small
  \centering
  \includegraphics[width=\linewidth]{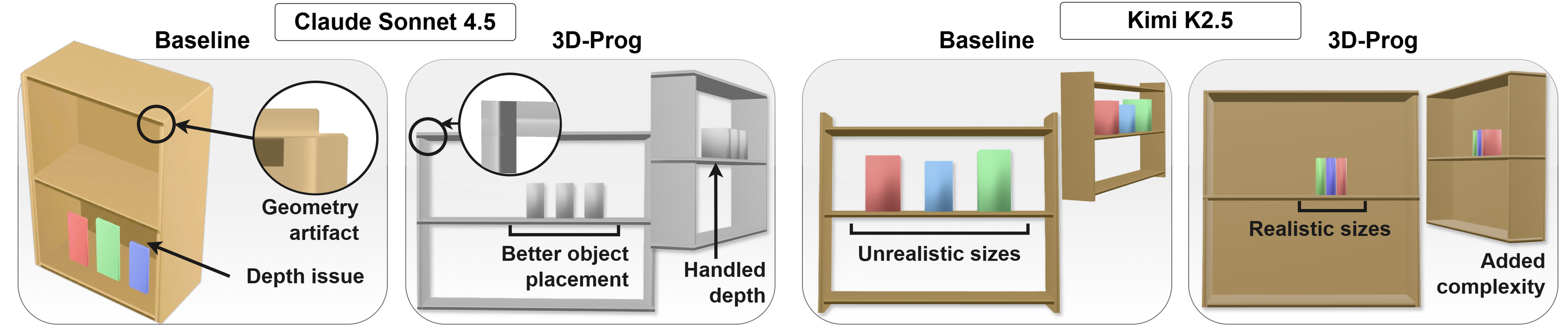}
  \caption{Qualitative results of 3D-Prog with Claude and Kimi backbones vs Baselines on Compositional Reasoning. Prompt: ``Generate a simple bookshelf with 3 books.''}
  \label{fig:quali}
\end{figure*}
\subsection{Results and Analysis}
We evaluate 3D-Prog across the tasks introduced in~\cref{sec:tasks} against diverse VLM baselines.
Shown in~\cref{tab:task_results}, 3D-Prog consistently outperforms or matches both proprietary and open-source baselines across all benchmarks, demonstrating how 3D-Prog equips VLMs with 3D generative, understanding, and reasoning capabilities.

For generative tasks, 3D-Prog is able to simplify and complete both dense complex object scans and synthetic assets (e.g., $0.19$ Chamfer distance for synthetic data completion and $0.08$ for Scan2Sim), indicating high geometric fidelity.
The Scan2Sim task shows a critical point of model capabilities, as all baseline models proved incapable of generating valid USD files from real-world data. These failures stem from either (a) non-compliance with simplification instructions, leading to context window overflow during the ingestion of dense meshes, or (b) catastrophic geometric loss caused by the models' inability to reconcile real-world scan noise and topological inconsistencies.

3D-Prog further demonstrates strong performance on 3D understanding tasks. 
Object detection and instance segmentation performance nearly double, confirming significantly improved 3D spatial reasoning at both the scene and object levels.
Furthermore, user study results for object editing favor 3D-Prog, indicating capabilities that extend beyond pure geometric understanding.

3D-Prog exhibits robust spatial reasoning, evidenced by its articulation predictions. It correctly identifies the motion type in every case and predicts the motion axis with a median error of $3.5^\circ$ and $0.6^\circ$ with the Claude and Kimi backbones, whereas standalone 2D VLMs err by nearly $90^\circ$. In the compositional reasoning task, evaluators show a strong preference for the scenes generated by 3D-Prog, highlighting their superior geometric and visual quality. 
Assets generated by 3D-Prog exhibit clean geometry with no part or object-level intersections, representative shapes for each object class, and realistic scale, materials, and color, leading to visually coherent and physically plausible compositions. 
In contrast, baseline outputs often contain non-distinctive geometry, unrealistic element proportions, detached components, and frequent mesh intersections between separate parts, resulting in visually unstable or semantically inconsistent scenes.  
Qualitative comparisons are shown in~\cref{fig:quali}, with detailed analysis of evaluator feedback from both user studies provided in the Suppl. material (\cref{sec:user_study}).

\subsection{Ablation Study}
\label{subsec:ablation}
We conduct ablation studies to examine the role of the CCF and TAF across various asset categories and both understanding and generative tasks, as shown in \cref{tab:ablation}. Specifically, we assess 3D-Prog across three distinct benchmarks: (a) instance segmentation, to evaluate fine-grained geometric and object-level comprehension; (b) object detection, to probe scene-level understanding; and (c) object completion, to demonstrate the model’s integrated capacity for geometric analysis and generative synthesis.
\begin{table}
    \scriptsize
    \setlength{\tabcolsep}{4pt}
    \centering
\caption{Ablation results on diverse tasks spanning generation and understanding.}
\begin{tabular}{l cccc}
\toprule
 & \textbf{3D-Prog} & \textbf{Claude} & w/o \textbf{CCF} & w/o \textbf{TAF} \\
 \midrule
Instance Segmentation (mAP $\uparrow$)& \textbf{0.42} & 0.21 & 0.30 & 0.29 \\
Object Detection (mAP $\uparrow$) &\textbf{0.38} & 0.04 & 0.01 & 0.11\\
Object Completion (Chamfer $\downarrow$) & \textbf{0.19} & 0.42 & 0.33 & 0.58\\
\bottomrule
\end{tabular}
\label{tab:ablation}
\end{table}
The results confirm that CCF and TAF are complementary. CCF provides the geometric grounding essential for metric reasoning---without it, completion Chamfer distance worsens from 0.19 to 0.33 and detection mAP collapses from 0.38 to 0.01. TAF enables iterative error correction; its removal eliminates the refinement loop, causing a 3$\times$ degradation (0.19 to 0.58). This disentangles the contribution: CCF grounds spatial understanding, while TAF provides the iterative scaffold for complex tasks. Specifically, omitting CCF restricts the visual prompt to a single orthographic view without coordinate framing, introducing inaccurate initial coordinate estimates that the VLM cannot rectify.
Similarly, omitting TAF removes the task-adaptive analysis that drives the iterative refinement loop, reducing the system to a single generation pass. Without iterative analysis, initial noisy outputs remain uncorrected, resulting in a sharp performance decline.

\begin{table}[t]
\centering
\scriptsize
\setlength{\tabcolsep}{3pt}
\caption{\textbf{Robustness to geometric perturbations.} Chamfer distance ($\downarrow$) on an object completion asset when the input and its ground truth are translated along $Y$ or rotated about $Z$; 3D-Prog (Claude backbone) vs.\ the Claude Sonnet 4.5 baseline. The average Chamfer distance is calculated over the six perturbed settings.}
\begin{tabular}{l c ccc ccc c}
\toprule
& \textbf{Original} & \multicolumn{3}{c}{\textbf{Translated} (Y)} & \multicolumn{3}{c}{\textbf{Rotated} (Z)} & \textbf{Avg.} \\
\cmidrule(lr){3-5} \cmidrule(lr){6-8}
\textbf{Method} & & +10cm & +20cm & +30cm & +15° & +30° & +45° & \\
\midrule
3D-Prog & \textbf{0.02} & \textbf{0.02} & \textbf{0.02} & \textbf{0.02} & \textbf{0.02} & \textbf{0.07} & 0.18 & \textbf{0.05} \\
Claude  & \textbf{0.02} & 0.38 & 0.44 & 0.11 & 0.33 & 0.18 & \textbf{0.17} & 0.27 \\
\bottomrule
\end{tabular}
\label{tab:robustness_wide}
\end{table}

\textbf{Robustness to geometric perturbations.} To test the stability of CCF grounding, we translate and rotate the input of the object completion task (transforming the ground truth accordingly) and compare against the Claude baseline in~\cref{tab:robustness_wide}. Both methods are equally accurate on the unperturbed input, but 3D-Prog holds a constant $0.02$ Chamfer distance under every translation and the $15^\circ$ rotation, and $0.07$ at $30^\circ$, whereas the baseline degrades by up to $22\times$ and fluctuates unpredictably between perturbation magnitudes. Only at $45^\circ$ do both methods degrade similarly.

\begin{figure*}[tb]
    \centering
\includegraphics[width=\linewidth]{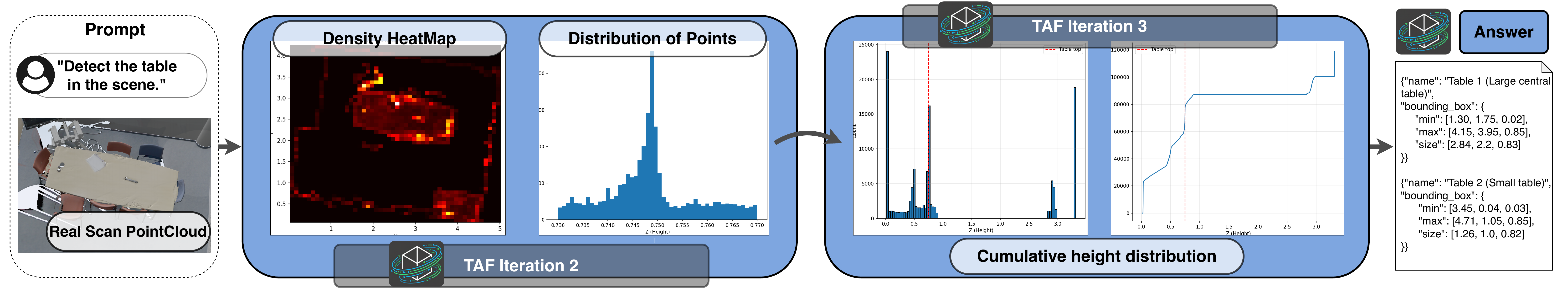}
    \caption{\textbf{TAF across iterations} for table detection in a scanned scene. Early iterations produce two cues: a density heatmap of points at typical table height (candidate footprints) and a height distribution (expected vertical extent). The next iteration restricts the analysis to the extracted footprint and uses the height distribution and its cumulative counterpart for noise rejection, yielding the final 3D bounding box.}
    \label{fig:task_adaptive}
\end{figure*}
\subsection{3D Programming Across VLM Families}
To evaluate how broadly 3D-Prog can transform 2D VLMs into 3D programmers, we benchmark it on various closed-source and open-source 2D VLMs. As shown in~\cref{tab:vlm_agnostic}, 3D-Prog yields consistent improvements across diverse VLM families, both closed-source and open-source.

\begin{table}[t]
\centering
\scriptsize
\setlength{\tabcolsep}{3pt}
\caption{\textbf{3D-Prog across various VLMs} on object completion, with state-of-the-art open- and closed-source 2D VLMs as backbones.}
\begin{tabular}{l c cccc}
\toprule
& \textbf{Baseline} & \multicolumn{4}{c}{\textbf{3D-Prog (Ours)}} \\
\cmidrule(lr){2-2} \cmidrule(lr){3-6}
\textbf{Metric} & Sonnet 4.5 & \makecell{GPT-5.1 \\ \includegraphics[height=8pt]{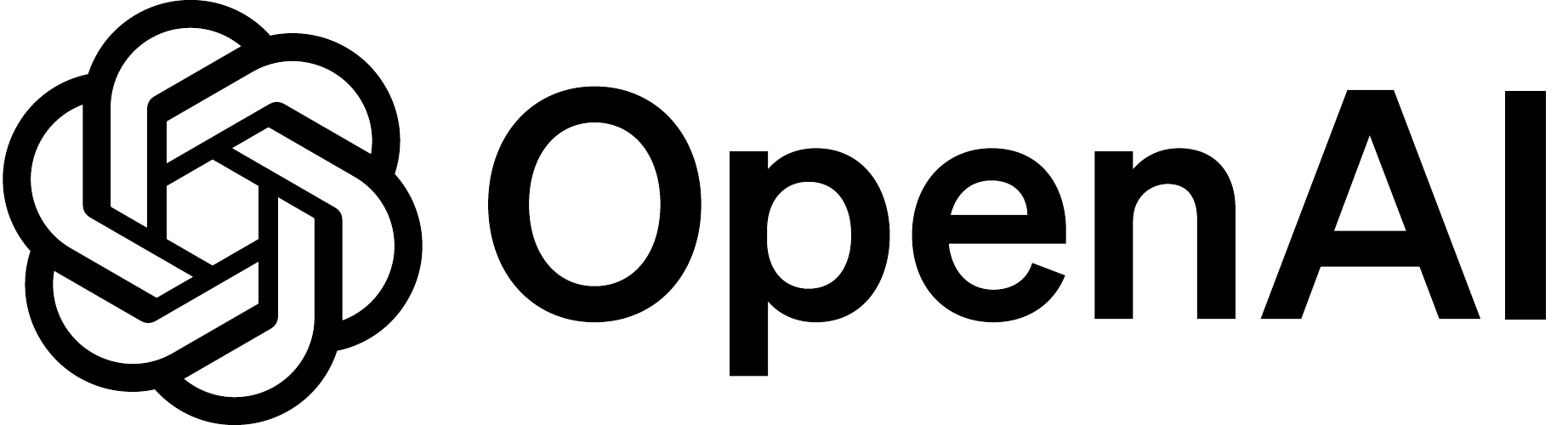}} & \makecell{Gemini 3 Pro \\ \includegraphics[height=8pt]{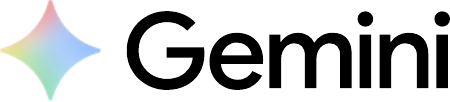}} & \makecell{Sonnet 4.5 \\ \includegraphics[height=8pt]{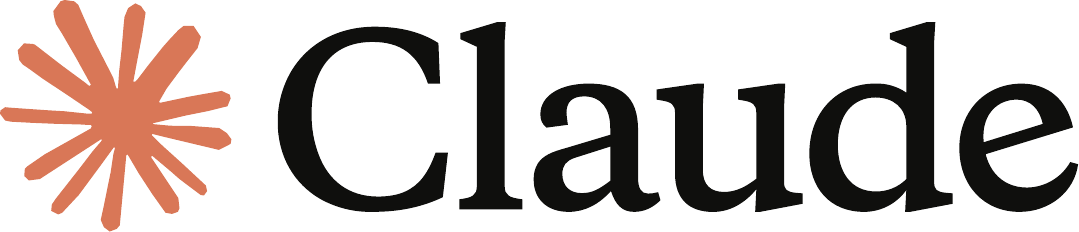}} & \makecell{Kimi K2.5\\ \includegraphics[height=8pt]{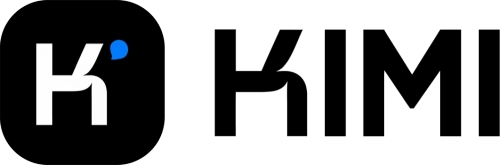}} \\
\midrule
Chamfer $\downarrow$ & 0.42 & 0.23 & 0.25 & 0.19 & \textbf{0.16} \\
\bottomrule
\end{tabular}
\label{tab:vlm_agnostic}
\end{table}

\textbf{Open-source vs closed-source VLMs for 3D programming.}
Our benchmark of open-source VLMs includes Kimi K2.5, Llama 4 \cite{llama4}, and Qwen3-VL-30B-A3B-Instruct~\cite{qwen3vltechnicalreport}. Of these models, only Kimi K2.5 demonstrates the required Python code generation capabilities and adaptability to the USD format. The failure of other models to produce valid outputs or process 3D data highlights a gap in open-source models for geometry-aware code generation. Although open-source VLMs like Qwen and Llama have sufficient visual understanding, they lack adequate coding/programming expertise for these tasks. However, the promising results from Kimi K2.5 suggest that high-performance 3D programming is becoming increasingly attainable within open-source frameworks.

\begin{table}[t]
\centering
\small
\setlength{\tabcolsep}{4pt}
\caption{\textbf{Comparisons to 3D VLMs.} 3D-Prog outperforms GaussianVLM~\cite{gaussianvlm} on both SQA3D and Nr3D.}
\begin{tabular}{l ccc}
\toprule
& \textbf{Claude} & \textbf{GaussianVLM} & \textbf{3D-Prog} \\
\midrule
SQA3D (Accuracy $\uparrow$) & 30\% & 50\% & \textbf{60\%} \\
Nr3D (METEOR $\uparrow$) & 0.16 & 0.22 & \textbf{0.23} \\
\bottomrule
\end{tabular}
\label{tab:3dvlms}
\end{table}

\subsection{Comparisons to 3D VLMs}

We evaluate the performance of 3D-Prog against a state-of-the-art 3D VLM~\cite{gaussianvlm} across two primary dimensions: scene-level and object-level 3D understanding. 
For scene-level evaluation, we assess the model’s ability to perform situated reasoning using the SQA3D benchmark \cite{ma2022sqa3d}, which requires interpreting complex environments from a specified egocentric perspective (e.g., “I am sitting on the bed facing the chair. What is to my direct left?”).
Object-level capabilities are evaluated on object captioning, where the model is provided with object bounding boxes and tasked with generating captions for the corresponding objects using the Nr3D benchmark \cite{achlioptas2020referit_3d}.
Performance is measured using accuracy for SQA3D and the METEOR \cite{banerjee2005meteor} captioning metric for Nr3D.
We compare 3D-Prog with the Claude 2D VLM baseline and GaussianVLM, a generalist 3D VLM \cite{gaussianvlm}. Results in \cref{tab:3dvlms} show that our training-free approach outperforms the state-of-the-art 3D VLM specifically trained on these datasets and tasks.

\subsection{Discussion of Experiments}
The analysis of 3D-Prog reveals that CCF is essential for tasks fundamentally rooted in geometric reasoning, e.g., object completion or compositional generation. In these settings, resolving gauge ambiguity and providing spatial grounding enable 3D-Prog to converge to a solution consistently over multiple iterations.
As expected, for tasks that primarily require appearance changes without geometric adaptations (e.g., simple object editing), the baseline model can achieve a reasonable success score by mapping the task to the 2D domain. Nevertheless, even in these scenarios, our approach outperforms the baseline, showcasing that additional feedback is consistently beneficial.

In a detailed analysis, we observe that visual feedback produced by the TAF module shows a strong degree of specialization tailored to the specific user query and task category. 
TAF demonstrates the ability to construct specialized visual feedback, generate complex NumPy code, and employ classical ML algorithms such as clustering for segmentation tasks.
Furthermore, TAF exhibits an inherently evolutionary nature: across successive iterations, its outputs become increasingly fine-grained by leveraging insights from preceding passes. 
This evolution from broad analysis to focused refinement is exemplified in the table detection task (\cref{fig:task_adaptive}). 
The first iteration generates general cues based on universal rules, while in subsequent iterations, the model shifts to focused asset-specific region analysis. Additional qualitative examples of TAF and a comprehensive failure analysis for both 3D-Prog and the baseline are provided in the Supplementary Material (\cref{sec:taf_supp}, \ref{sec:error_supp}).

Across evaluation tasks, 3D-Prog converges in 7--20 iterations (mean: 11), with each iteration taking approximately 22s on average. Total runtime ranges from around 2 minutes for articulation and QA tasks to 8 minutes for segmentation. VLM inference dominates compute, followed by code execution and rendering. A detailed per-task breakdown is provided in the Supplementary Material.
\section{Conclusion}
\label{sec:conclusion}
In this work, we presented 3D-Prog, a generalizable approach that enables 2D VLMs to perform 3D tasks, including understanding, reasoning, editing, composition, and generation. 
3D-Prog achieves this through CCF, which enables grounding and 3D perception in 2D VLMs, and TAF, which encompasses a flexible scratch-space for task-specific feedback.
Experiments further showcased that 3D-Prog is a powerful generalist framework that brings the capabilities of current 2D VLMs to free-form 3D programming, unlocking the potential of VLMs for 3D tasks.
We demonstrated the versatility and generalizability of 3D-Prog across a diverse set of tasks on both real and synthetic assets, where it consistently and reliably produces high-quality 3D outputs and achieves precise 3D understanding.
\paragraph{Limitations.}
Although powerful, 3D-Prog is not without limitations. 
First, by design, the performance of 3D-Prog depends on the capabilities of the underlying VLMs, 
which still struggle with very complex reasoning tasks or fine-grained details. 
Second, 3D-Prog relies on a judge module that evaluates convergence rather than correctness. While 100\% motion-type accuracy on articulation suggests the iterative loop is self-correcting, the judge may occasionally accept suboptimal results. We mitigate this through conservative iteration limits and visual comparison feedback.
Lastly, 3D-Prog's ability to process 3D assets relies on the CCF, which is limited in scene size and detail by the VLM context.
However, with the fast progress in open and closed VLMs, our method has a clear path towards scaling beyond its current performance.

\section*{Acknowledgments}
This research was partially funded by the Ministry of Education and Science of Bulgaria (support for INSAIT, part of the Bulgarian National Roadmap for Research Infrastructure).

{
    \small
    \bibliographystyle{ieeenat_fullname}
    \bibliography{main}
}

\clearpage
\setcounter{section}{0}
\setcounter{table}{0}
\setcounter{figure}{0}
\renewcommand{\thesection}{\Alph{section}}
\renewcommand{\thetable}{S\arabic{table}}
\renewcommand{\thefigure}{S\arabic{figure}}
\renewcommand{\theHsection}{S\Alph{section}}
\renewcommand{\theHtable}{S\arabic{table}}
\renewcommand{\theHfigure}{S\arabic{figure}}
\section*{Supplementary Material}

We provide an extended analysis and implementation details to support the main manuscript. The supplementary material is organized as follows: Sec.~\ref{sec:tasks_suppl} provides a detailed task taxonomy and dataset licenses. Sec.~\ref{sec:runtime} reports per-task runtime statistics. Sec.~\ref{sec:user_study} presents the methodology of our user study. We then provide a detailed TAF analysis (Sec.~\ref{sec:taf_supp}), followed by a qualitative error analysis (Sec.~\ref{sec:error_supp}) to investigate current limitations. We further expand the scope of our evaluation with qualitative results in Sec.~\ref{sec:qualitative}. We conclude by detailing the specific design choices (Sec.~\ref{sec:design_supp}), including the core algorithm and full prompt specifications.

\section{Task Overview}
\label{sec:tasks_suppl}
We demonstrate 3D-Prog across nine tasks spanning generation, understanding, and reasoning. Tab.~\ref{tab:tasks} provides a detailed breakdown of each task, its description, and input asset types.

\begin{table*}[t]
\centering
\scriptsize
\renewcommand{\arraystretch}{1.4}
\setlength{\tabcolsep}{5pt}
\caption{Summary of 3D programming tasks we demonstrate, spanning generation, understanding, and reasoning over 3D assets.}
\begin{tabular}{@{}c@{\hskip 4mm}p{3.3cm}@{\hskip 4mm}p{5.7cm}@{\hskip 2mm}p{1.9cm}@{}}
\toprule
& \textbf{Task} & \textbf{Description} & \textbf{Input Asset} \\
\midrule
\multirow{2}{*}[-0.25ex]{\rotatebox{90}{Generation}}
& Object Completion
& Complete partial object geometry into complete shapes.
& \makecell[tl]{3D object\\synthetic~\cite{openusd_kitchen_set,deitke2023objaverse}} \\[2pt]
\cline{2-4}
& Scan2Sim
& Convert a real-world scan into a simplified simulation-ready asset.
& \makecell[tl]{3D object\\scan~\cite{Halacheva2024HolisticUO}} \\[2pt]
\midrule
\multirow{5}{*}[-8ex]{\rotatebox{90}{Understanding}}
& \makecell[tl]{Object Detection\\\textcolor{gray}{(Spatial)}}
& Detect and localize objects with bounding boxes for perception or manipulation tasks.
& \makecell[tl]{3D scene scan/\\synthetic~\cite{openusd_kitchen_set,yeshwanthliu2023scannetppSup}} \\[2pt]
\cline{2-4}
& \makecell[tl]{Object Part Segmentation\\\textcolor{gray}{(Geometric)}}
& Segment objects into parts or sub-components with semantic and/or instance labels.
& \makecell[tl]{3D object\\synthetic~\cite{openusd_kitchen_set,deitke2023objaverse}} \\[2pt]
\cline{2-4}
& \makecell[tl]{Object Editing\\\textcolor{gray}{(Visual)}}
& Modify object textures and materials; adjust object pose or placement at scene level.
& \makecell[tl]{3D object\\synthetic~\cite{openusd_kitchen_set,deitke2023objaverse}} \\[2pt]
\cline{2-4}
& \makecell[tl]{Situated QA\\\textcolor{gray}{(Contextual)}}
& Spatial reasoning from a situated perspective.
& \makecell[tl]{3D scene\\scan~\cite{ma2022sqa3d}} \\[2pt]
\cline{2-4}
& \makecell[tl]{Object Captioning\\\textcolor{gray}{(Semantic-Spatial)}}
& Given a 3D scene scan and a 3D bounding box of an object within it, caption the object.
& \makecell[tl]{3D scene\\scan~\cite{achlioptas2020referit_3d}} \\[2pt]
\midrule
\multirow{2}{*}[-3ex]{\rotatebox{90}{Reasoning}}
& Compositional Reasoning
& Build a coherent scene from a given set of objects by reasoning about spatial and semantic relationships.
& \makecell[tl]{Text prompt\\(opt.\ 3D objects)} \\[2pt]
\cline{2-4}
& Articulation Prediction
& Given a 3D scene and a point marking a movable component, the model predicts the motion axis of the marked part.
& \makecell[tl]{3D scene\\scan~\cite{Halacheva2024HolisticUO}} \\[2pt]
\bottomrule
\end{tabular}
\label{tab:tasks}
\end{table*}

\paragraph{Dataset Licenses and Terms of Use.}
We report the licenses and terms of use for the datasets used in our evaluation in \cref{tab:dataset-licenses} and adhere strictly to each. All datasets are publicly available. We draw attention to ScanNet~\cite{Dai2017ScanNetR3}\footnote{Available at \url{https://kaldir.vc.in.tum.de/scannet/ScanNet_TOS.pdf} (last accessed: 27 August 2026).} and ScanNet++~\cite{yeshwanthliu2023scannetppSup}\footnote{Available at \url{https://kaldir.vc.in.tum.de/scannetpp/static/scannetpp-terms-of-use.pdf} (last accessed: 27 August 2026).} being subject to custom terms of use that require separate inspection, and to the Pixar Kitchen Set, which is distributed under Pixar's non-commercial USD asset license.

\begin{table}[ht]
\centering
\scriptsize
\setlength{\tabcolsep}{4pt}
\caption{Licenses and terms of use for datasets employed in our study.}
\begin{tabular}{l l}
\toprule
\textbf{Dataset} & \textbf{License / Terms of Use} \\
\midrule
SQA3D~\cite{ma2022sqa3d} & Apache 2.0 \\
Nr3D~\cite{achlioptas2020referit_3d} & MIT \\
ScanNet~\cite{Dai2017ScanNetR3} & ScanNet Terms of Use \\
ScanNet++~\cite{yeshwanthliu2023scannetppSup} & ScanNet++ Terms of Use \\
Articulate3D~\cite{Halacheva2024HolisticUO} & CC BY 4.0 \\
Objaverse-XL~\cite{deitke2023objaverse} & ODC-By 1.0 (per-object licenses vary) \\
Pixar Kitchen Set~\cite{openusd_kitchen_set} & Pixar non-commercial asset license \\
\bottomrule
\end{tabular}
\label{tab:dataset-licenses}
\end{table}

\section{Runtime Analysis}
\label{sec:runtime}

We report runtime statistics for 3D-Prog across select evaluation tasks in Tab.~\ref{tab:runtime}. Measurements were collected on an NVIDIA RTX 5090 GPU with 32 GB of memory.

\begin{table}[ht]
\centering
\scriptsize
\setlength{\tabcolsep}{4pt}
\caption{\textbf{Runtime analysis.} Total time in seconds. Per-iteration time includes VLM inference ($\sim$60\%), code execution ($\sim$25\%), and rendering ($\sim$15\%). Tasks requiring finer geometric reasoning (segmentation) use more iterations than semantic tasks (QA, captioning).}
\begin{tabular}{l ccc}
\toprule
\textbf{Task} & \textbf{Avg. Iter.} & \textbf{Total (s)} & \textbf{Per Iter. (s)} \\
\midrule
Object Completion & 12.7 & 265 & 21 \\
Object Detection & 12.0 & 287 & 24 \\
Semantic Seg. & 19.2 & 446 & 23 \\
Object Captioning & 7.4 & 216 & 29 \\
Situated QA & 9.8 & 195 & 20 \\
Articulation & 7.3 & 125 & 17 \\
\midrule
\textbf{Average} & \textbf{11.4} & \textbf{256} & \textbf{22} \\
\bottomrule
\end{tabular}
\label{tab:runtime}
\end{table}

\section{User Study}
\label{sec:user_study}
User studies with four independent evaluators assessed 3D-Prog's performance on object editing and compositional reasoning tasks.
Evaluators scored each VLM-generated 3D asset from 0 to 5, following five predefined task-specific success criteria. 
For each instance, the evaluators were provided with the user prompt and the 3D visualization of the generated asset in a 3D USD visualizer.
For object editing, the input 3D asset was also provided to check for geometry preservation.
The final task performance was the average score from the four evaluators.

Compositional reasoning criteria cover semantics, orientation, color/material accuracy, structure, and composition (detailed in Tab.~\ref{tab:criteria_compos_reas}). Object editing criteria include geometry preservation, material accuracy, realism of physical properties, scope accuracy, and visual realism (detailed in Tab.~\ref{tab:criteria_edit}).

Analysis of the evaluators' scores shows that in compositional reasoning, the baseline suffered from mesh intersections, incorrect orientations, and indistinguishable semantics, particularly when handling non-cubic primitives. 
Its success was limited to color and material attributes. 
In contrast, 3D-Prog achieved consistently high performance across all criteria. 
Regarding object editing, evaluators reported comparable visual quality between methods; however, the baseline consistently failed to execute physical property corrections.

\begin{table}[t]
\centering
\scriptsize
\setlength{\tabcolsep}{5pt}
\caption{Success criteria for compositional reasoning.}
\begin{tabular}{l p{0.72\linewidth}}
\toprule
\textbf{Criterion} & \textbf{Description} \\
\midrule
\textbf{Semantics} & The overall compositional structure and scene semantics must exactly match task instructions from multi-angle inspection. \\
\midrule
\makecell[tl]{\textbf{Orientation}\\ \textbf{\& Alignment}} & Objects with implied directionality must be oriented realistically and consistently with natural placement or user requirements. \\
\midrule
\makecell[tl]{\textbf{Color}\\ \textbf{\& Material}} & Objects must have the correct specified colors. \\
\midrule
\textbf{Structure} & No unrealistic overlapping or intersection of objects unless explicitly required by the task. \\
\midrule
\textbf{Composition} & Objects must maintain correct compositional relationships (e.g., stacked/contained objects must not float or clip). \\
\bottomrule
\end{tabular}
\label{tab:criteria_compos_reas}
\end{table}

\begin{table}[t]
\centering
\scriptsize
\setlength{\tabcolsep}{5pt}
\caption{Success criteria for object editing.}
\begin{tabular}{l p{0.72\linewidth}}
\toprule
\textbf{Criterion} & \textbf{Description} \\
\midrule
\makecell[tl]{\textbf{Geometry}\\ \textbf{Preservation}} & The object's underlying geometry must remain exactly the same as the original (incl.\ proportions, rotations, translations, etc.). \\
\midrule
\makecell[tl]{\textbf{Material}\\ \textbf{Accuracy}} & The edited material must visually match the intended type. \\
\midrule
\makecell[tl]{\textbf{Physical}\\ \textbf{Properties}} & Roughness, density, metallic, specular properties, etc.\ must be correct. \\
\midrule
\makecell[tl]{\textbf{Scope}\\ \textbf{Accuracy}} & The edit is correctly implemented over the specified scope, regardless of whether that scope is the entire object or a localized part. \\
\midrule
\makecell[tl]{\textbf{Visual}\\ \textbf{Realism}} & The object must behave visually in a way consistent with the intended material (e.g., proper metallic reflectivity, ceramic glossiness). Highlights and reflections must behave correctly. \\
\bottomrule
\end{tabular}
\label{tab:criteria_edit}
\end{table}

\section{TAF Analysis}
\label{sec:taf_supp}
In this section, we conduct a thorough analysis of the TAF component, with Sec.~\ref{subsec:taf_analysis} providing an inspection of the TAF feedback mechanism and Sec.~\ref{sec:iter_case_study} giving an in-depth walkthrough of an example rollout of 3D-Prog on the semantic segmentation task with a 3-step iterative refinement.
\subsection{Structure of TAF Outputs}
\label{subsec:taf_analysis}
TAF is a central component of 3D-Prog, enabling it to (a) extract semantic knowledge from VLMs, (b) reason about task execution in an iterative closed loop, thereby improving geometric fidelity and performance with respect to the number of iterations, and (c) address open-ended tasks. 

We conducted a manual inspection of the 3D-Prog logs to analyze the capabilities of the TAF. The typical format of TAF follows the structure below, where each component exhibits distinct capabilities.
\begin{itemize}
    \item \textbf{Analysis}: As instructed in the prompt, the VLMs perform both visual and structural (geometric) analysis for the given tasks, largely independent of the task specifics. This includes (a) inferring the axis of the 3D asset, (b) visually estimating shapes, sizes, and semantically identifying the 3D assets, and (c) inspecting the bounds of the asset.
    \item \textbf{3D Programming}: Uses the results of the analysis to carry out highly task-specific operations. Several steps may be carried out to gather more details about the asset, including (a) geometric filtering to remove noise, (b) finding the region of interest and visually zooming in, (c) PCA of the asset to estimate the canonical pose and the axis of symmetry, and (d) clustering objects and analyzing the individual clusters.
    \item \textbf{Reasoning}: Based on the analysis and the 3D programming outputs, the model draws conclusions that guide the next steps of the iteration. This typically involves (a) finding flaws in the analysis and the 3D programming (e.g., DBSCAN clusters are too coarse); (b) providing candidate solutions to resolve these flaws (reducing the DBSCAN $\epsilon$ for fine-grained clusters).   
    \item \textbf{Visual Validation}: The final stage involves visually inspecting the output to verify its correctness and validating the geometry and semantics of the asset.  
\end{itemize}

\subsection{Iterative Refinement Case Study: Semantic Segmentation}
\label{sec:iter_case_study}

\begin{figure*}[tb]
    \centering
    \includegraphics[width=\linewidth]{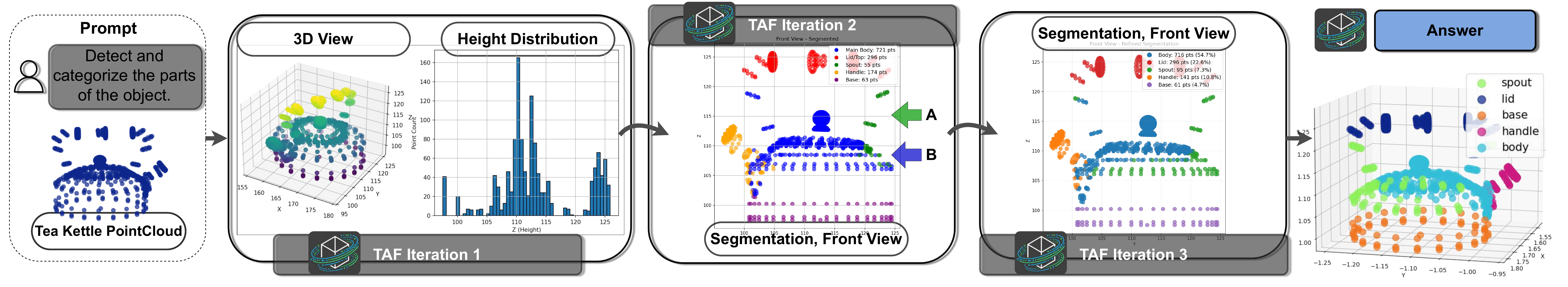}
    \caption{\textbf{TAF Visualizations Across Iterations.} The first iteration focuses on initial analysis for object understanding and label set prediction. In the second iteration, the TAF visualizes the initial segmentation of the tea kettle. We observe that the first segmentation exhibits over-segmentation (A), as well as under-segmentation issues (B). Those issues are detected by the VLM through the TAF and refined in the third iteration.}
    \label{fig:taf_iter}
\end{figure*}

We perform a case study investigating the TAF visualizations across different 3D-Prog iterations.
For the case study, we consider a semantic segmentation task on a synthetic tea kettle instance provided as a point cloud, as shown in Fig.~\ref{fig:taf_iter}.

\paragraph{Iteration 1: Initial Analysis and Hypothesis Formulation.}

The first visual program checked the asset's up-axis and traversed the stage hierarchy, establishing that the asset contains a single mesh.
Subsequently, the VLM calculated the bounding box and formulated an initial hypothesis that the semantic parts of the kettle are \textit{Main Body}, \textit{Lid}, \textit{Spout}, \textit{Handle}, and \textit{Kettle Base}.
The generated TAF was a 6-panel plot showing front, right, and top views, two 3D views, and a height distribution histogram. 

The generated multi-view point-cloud analysis visualization was utilized to confirm the initial hypothesis regarding the presence of the five target components and to establish the basis for segmentation using $Z$ (height) and $R$ (radial distance from the $XY$ centroid) as primary features.
\paragraph{Iteration 2: Segmentation and Validation.}

The initial segmentation algorithm employed geometric filtering for the \textit{Base} and \textit{Lid} (top 20\% of $Z$-range). The middle section was separated into the \textit{Body} (points within the 70th percentile radial distance) and an outer region.

The outer points were subjected to DBSCAN clustering to distinguish the \textit{Spout} and \textit{Handle} based on size and vertical spread.
The second TAF, a 6-panel visualization with semantically colored points, was generated for visual inspection. 

\paragraph{Iteration 3: Refinement with Enhanced Spatial Heuristics.}

The visual inspection of the TAF showed under-segmentation on the \textit{Lid} and \textit{Main Body} parts, as well as over-segmentation on the \textit{Handle}. The VLM refined the initial segmentation algorithm with the following adjustments:

\begin{enumerate}
    \item Threshold Adjustment: Height thresholds for the \textit{Base} and \textit{Lid} were slightly narrowed (12\% and 18\%, respectively), and the radial threshold was increased to the 75th percentile to ensure the \textit{Handle} and \textit{Spout} were fully contained within the outer region.
    \item Refined Clustering: DBSCAN parameters were adjusted to \texttt{eps=2.5} and \texttt{min\_samples=3} for improved density sensitivity.
    \item Geometric Differentiation Heuristics: The critical refinement was the application of post-DBSCAN geometric heuristics to definitively classify each resulting cluster as either the \textit{Spout} or the \textit{Handle}. The \textit{Handle} was identified by its arc shape and vertical dominance, whereas the \textit{Spout} was identified by its relatively compact and linear protrusion.
\end{enumerate}

The legend of the analyzed TAF from iteration 2 also reveals a semantic labeling error---the segment labels were distributed incorrectly (e.g., the \textit{Handle} was called ``Lid'', the \textit{Spout} was labeled as ``Handle''). This inconsistency was obvious in the legend of the plot, but the VLM failed to attend to this non-geometric metadata and did not attempt to correct it, retaining the segment's misclassification.

The third TAF is another panel with different views of the semantically colored point cloud, which confirmed the successful geometric segmentation.
Crucially, while the geometric output was correct, the VLM did not self-correct the initial semantic labeling error, and the \textit{components remained incorrectly labeled}. 
Notably, this failure to self-correct the labeling is observed very rarely and may be linked to the complexity of the task and to the VLM hallucinations.

\section{Error Analysis}
\label{sec:error_supp}
For each of the runs, we manually inspect the breakdown of the error modes of 3D-Prog.
This analysis is essential to understand both the strengths of our method and its current limitations.
We categorize the primary error modes as follows:
\begin{itemize}
    \item \textbf{Incorrect Program}: The most common failure mode is the method producing erroneous code or invalid USD asset formats. This error mode is more common in the baseline system, often resulting in outputs that cannot be rendered. Iterative feedback significantly reduces these issues in 3D-Prog.
    \item \textbf{Insufficient Geometric Fidelity}: The final output sometimes lacks fine-grained geometric details. In some cases, 3D-Prog oversimplifies the geometry, behaving ``\textit{lazily}'' and failing to capture intricate structures. The independent judge helps reduce this issue by enforcing stronger critical feedback.
    \item \textbf{Trivial Mistakes by Hallucination}: These occur when there is a mismatch between the model’s analysis and its final output. In a few runs, 3D-Prog correctly reasons about the scene but fails to transfer the same findings into the generated output, mainly due to hallucinations in the VLMs, as illustrated by the example in Sec.~\ref{sec:iter_case_study}.
\end{itemize}

\section{Qualitative Results}
\label{sec:qualitative}
\begin{figure}[t]
    \centering
    \includegraphics[width=0.8\linewidth]{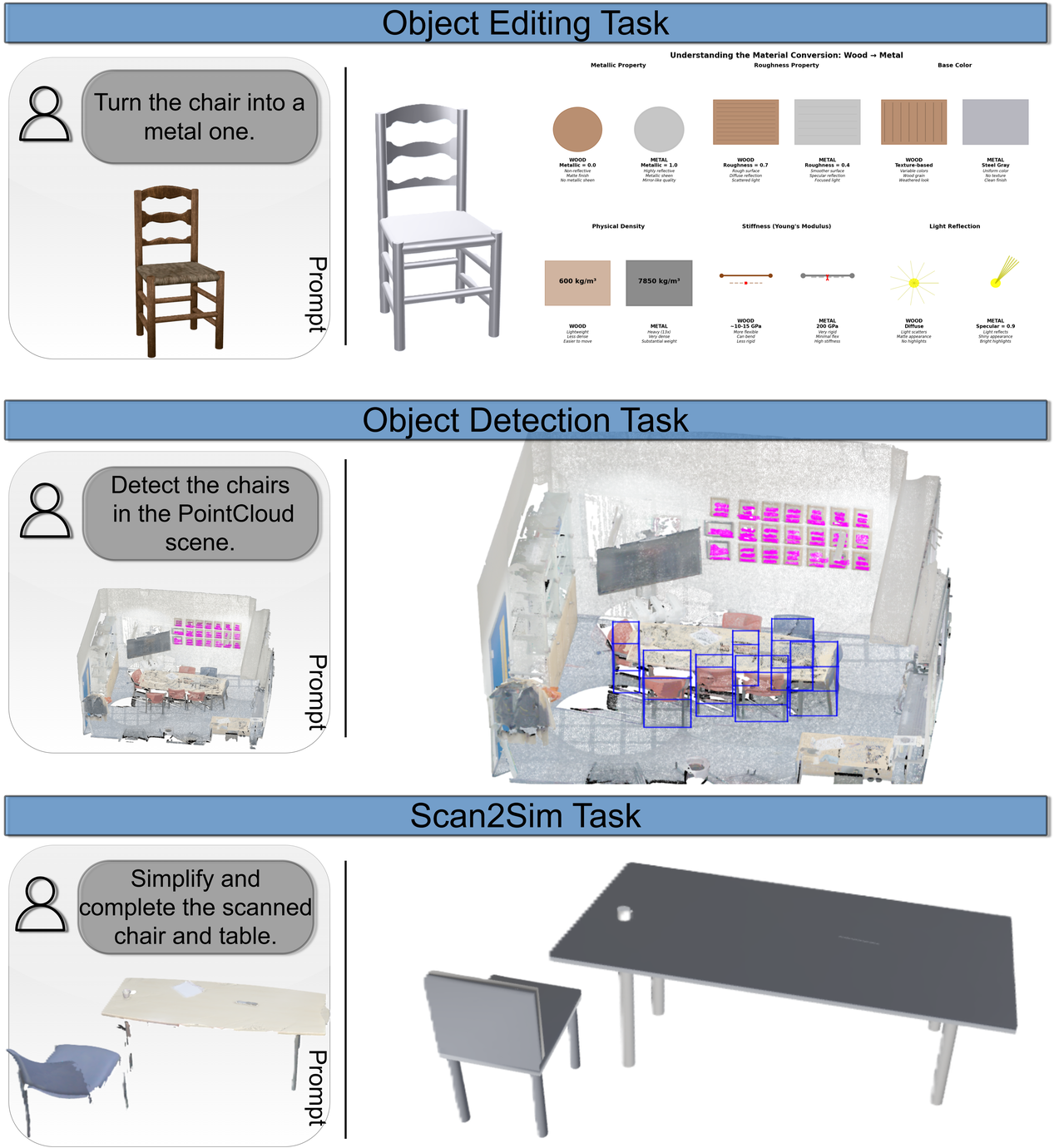}
    \caption{\textbf{Qualitative Results on 3D Object Editing, 3D Object Detection, and Scan2Sim Tasks.} For object editing, 3D-Prog generates a new USD asset with edited appearance and the corresponding physical properties. For object detection, the output is a list of bounding boxes in textual format, and in Scan2Sim the generated asset is the USD file of the simplified object.}
    \label{fig:quali_all}
\end{figure}
We present additional qualitative results in Fig.~\ref{fig:quali_all} to further demonstrate the capabilities of our proposed framework. In object editing, 3D-Prog successfully modifies both the appearance and physical characteristics. For object detection, we demonstrate the identification of chairs within a cluttered environment, effectively handling incomplete scans and multiple true positives. Finally, we highlight 3D-Prog's ability to simultaneously perform scan simplification and completion to produce simulation-ready assets across diverse object classes and sizes.

\subsection{3D-Prog for 3D Programming in the Wild}

\begin{figure}[tb]
    \centering
    \includegraphics[width=\linewidth]{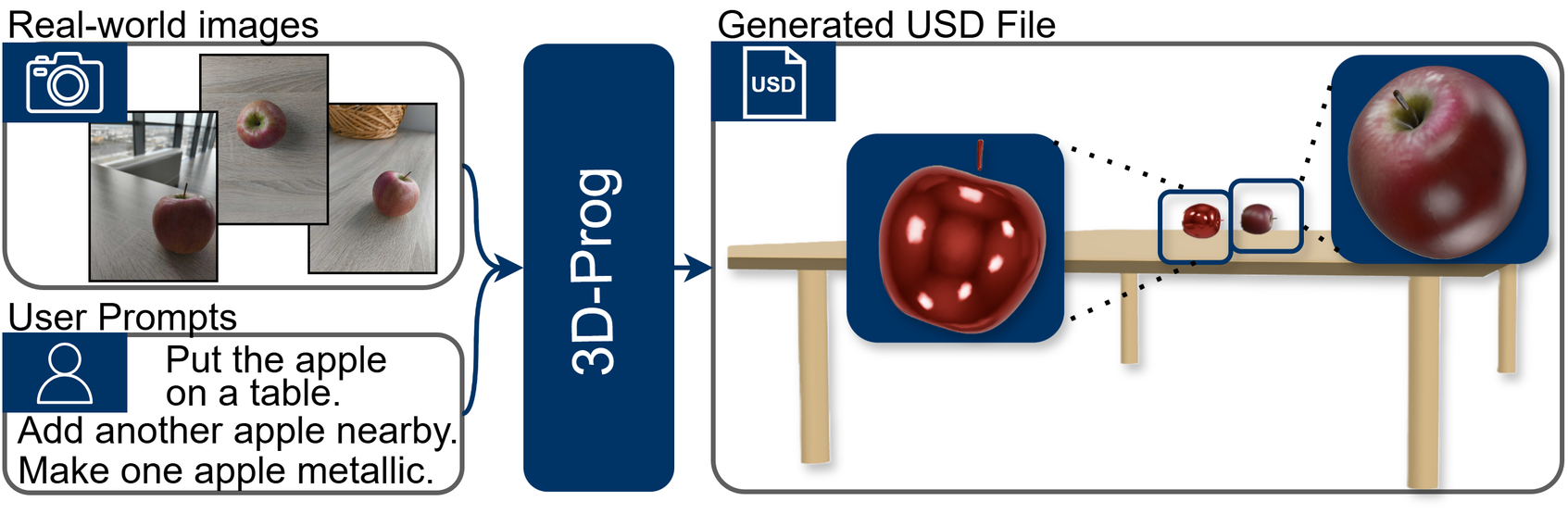}
    \caption{Zero-shot 3D-Prog results using self-captured RGB images.}
    \label{fig:real-images}
\end{figure}

We further validate 3D-Prog on self-captured real images taken in the wild with a phone camera. The first apple is reconstructed using SAM 3D \cite{chen2025sam}. 
Then we run 3D-Prog for various tasks, namely a) Generation and scene composition: to generate and place the table. The table needs to be of reasonable dimensions and texture. b) Object editing: to generate another apple and make its texture metallic. 
We show results in Fig.~\ref{fig:real-images}, where the table and the second apple are generated by 3D-Prog.

\section{Design Choices}
\label{sec:design_supp}
In this section, we detail the specific implementation strategies. First, we elaborate on the 3D-Prog algorithm (Sec.~\ref{sec:alg}), and in Sec.~\ref{sec:prompts} we provide the full prompts utilized within the framework.

\begin{figure*}[t]
  \centering
  \begin{subfigure}[b]{0.13\textwidth}
    \centering
    \includegraphics[width=\linewidth, trim={20 30 20 20}, clip]{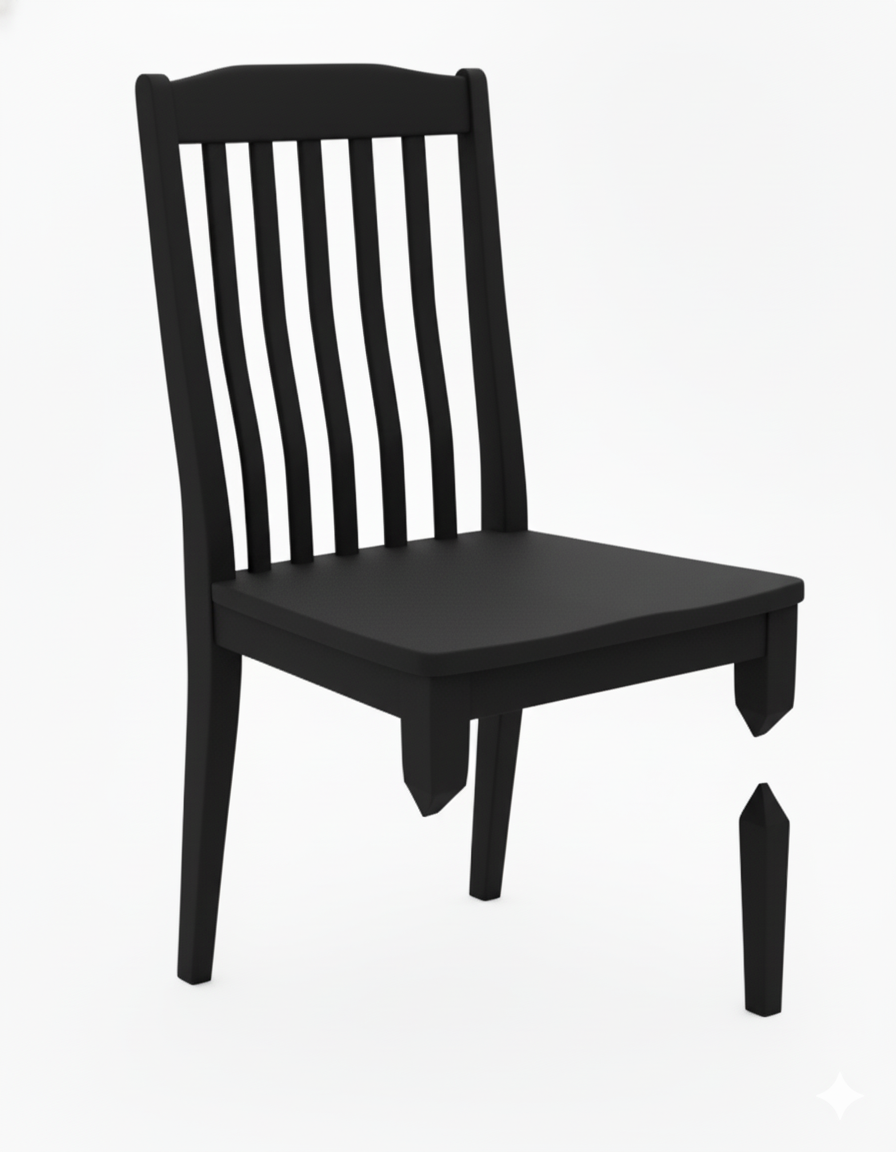}
    \caption{Input asset}
  \end{subfigure}\hfill
  \begin{subfigure}[b]{0.13\textwidth}
    \centering
    \includegraphics[width=\linewidth, trim={20 30 20 20}, clip]{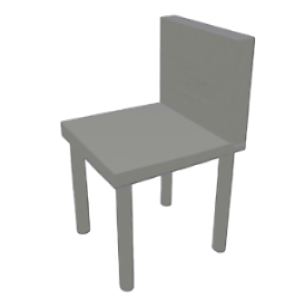}
    \caption{Generated asset}
  \end{subfigure}\hfill
  \begin{subfigure}[b]{0.34\textwidth}
    \centering
    \includegraphics[width=\linewidth]{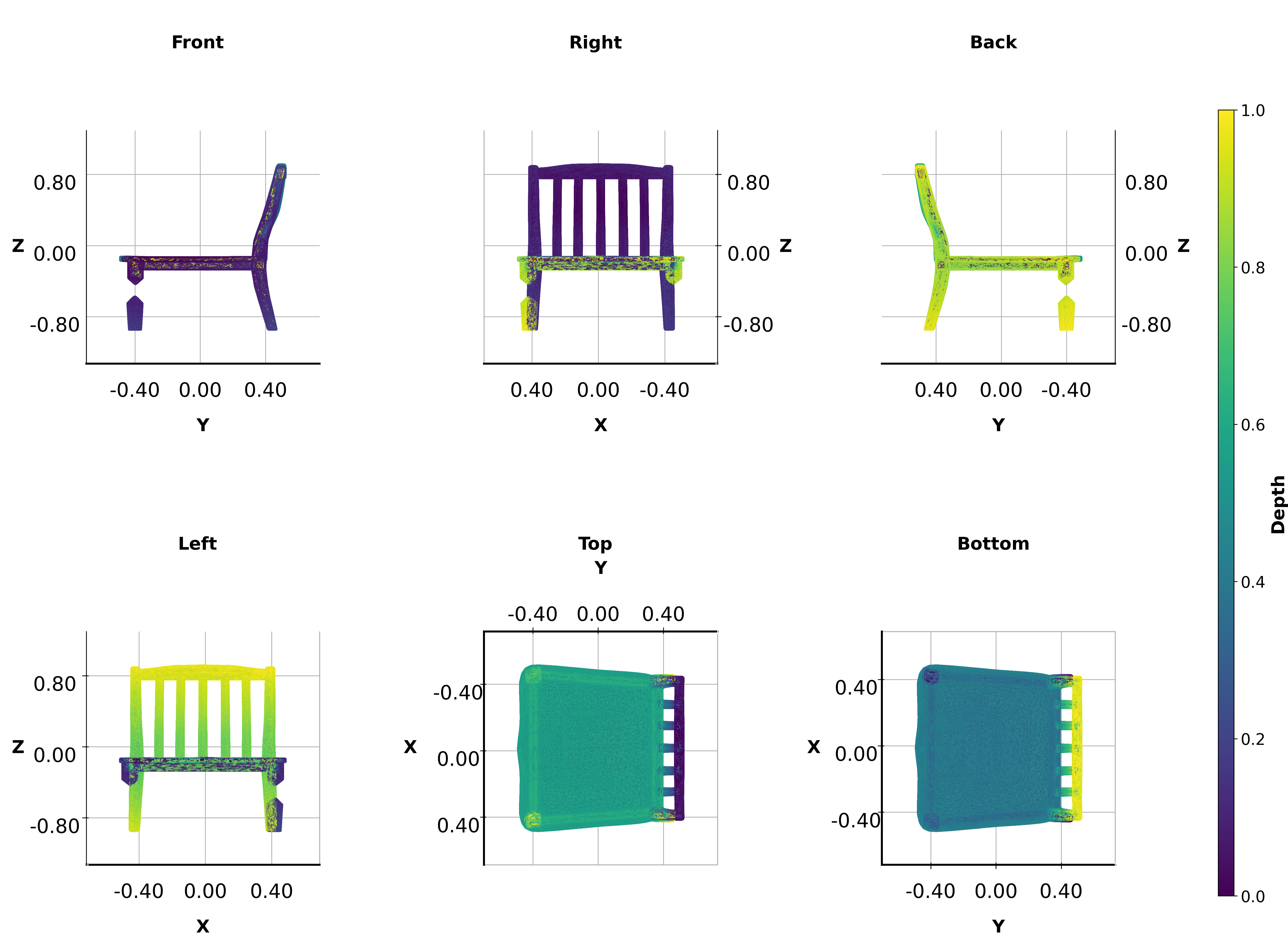}
    \caption{Initial visual prompt (iter.~1)}
    \label{fig:vis_prompt_1}
  \end{subfigure}\hfill
  \begin{subfigure}[b]{0.34\textwidth}
    \centering
    \includegraphics[width=\linewidth]{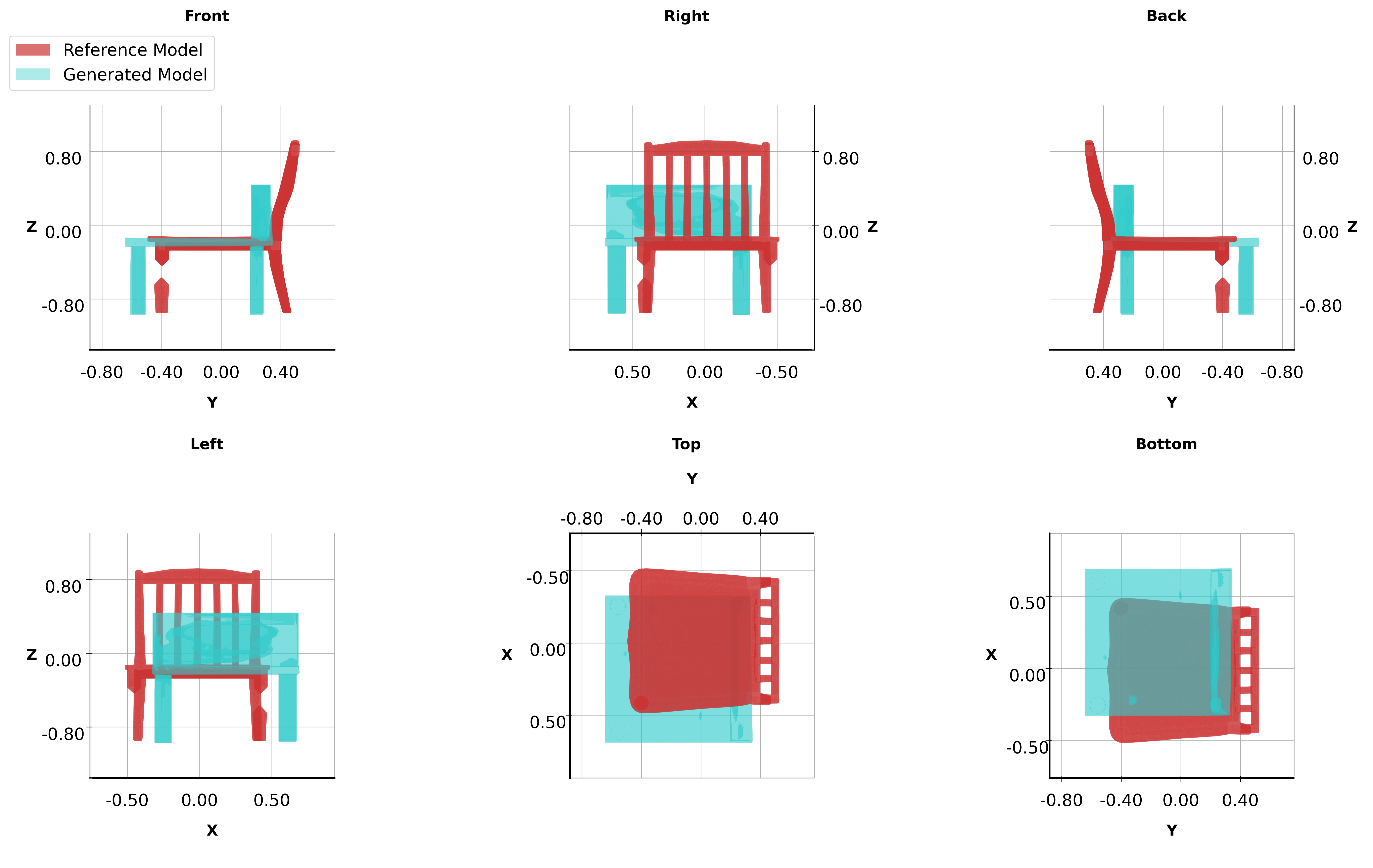}
    \caption{Comparative visual feedback (iter.~$>$1)}
    \label{fig:vis_prompt_2}
  \end{subfigure}
  \caption{\textbf{Visual prompt for 3D-Prog.} (a)~Input asset and (b)~the asset generated in iteration~1. (c)~The initial visual prompt after CCF: six orthographic views with depth-mapping. (d)~The comparative visual feedback used from the second iteration on: six orthographic views overlaying the current generated asset (cyan) on the reference asset (red, when available). The complete visual prompt for iterations $>$1 additionally contains the TAF image, which varies with the task and iteration.}
  \label{fig:visual_prompt}
\end{figure*}
\subsection{3D-Prog Algorithm}
\label{sec:alg}
In this section, we provide the complete algorithmic framework. Specifically, Algorithm~\ref{alg:visual_reasoning} details the Visual Reasoning for 3D Programming process. This algorithm formalizes how the system repeatedly uses multi-view visual feedback to:

\begin{algorithm}[tb]
\centering
\caption{Visual Reasoning for 3D Programming}
\label{alg:visual_reasoning}
\begin{tcolorbox}[colback=gray!5,colframe=gray!40,boxrule=0.4pt,
left=4pt,right=4pt,top=4pt,bottom=4pt]
\begin{algorithmic}[1]
\Require Reference model $\mathcal{R}$ (optional), Task specification $T$, VLM $\mathcal{V}$
\Ensure Generated USD model $\mathcal{M}$ satisfying task requirements
\State $\mathcal{I} \gets \text{RenderMultiView}(\mathcal{R})$ \Comment{6 orthographic views; $\emptyset$ if no $\mathcal{R}$}
\State $iteration \gets 0$;\; $feedback \gets \emptyset$
\Repeat
    \State $iteration \gets iteration + 1$
    \State $\mathcal{A} \gets \mathcal{V}.\text{AnalyzeGeometry}(\mathcal{I}, T, feedback)$ \Comment{Extract geometric features}
    \State $code \gets \mathcal{V}.\text{GenerateUSD}(\mathcal{A})$ \Comment{Generate manipulation code}
    \State $\mathcal{M} \gets \text{Execute}(code)$ \Comment{Create/modify USD geometry}
    \State $\mathcal{I}_{gen} \gets \text{RenderMultiView}(\mathcal{M})$;\; $\mathcal{C} \gets \mathcal{I}_{gen}$
    \If{$\mathcal{R}$ exists}
        \State $\mathcal{C} \gets \text{CompareVisual}(\mathcal{M}, \mathcal{R})$ \Comment{Overlay comparison}
        \State $feedback \gets \mathcal{V}.\text{AnalyzeDiscrepancy}(\mathcal{C})$
    \Else
        \State $feedback \gets \mathcal{V}.\text{EvaluateQuality}(\mathcal{I}_{gen})$
    \EndIf
    \State $approved \gets \text{Judge.Evaluate}(\mathcal{M}, T, \mathcal{C})$
\Until{$approved$ \textbf{or} $iteration \geq \text{MAX\_ITERATIONS}$}
\State \Return $\mathcal{M}$
\end{algorithmic}

\end{tcolorbox}
\end{algorithm}

\begin{enumerate}
    \item Analyze the target or generated geometry.
    \item Generate Universal Scene Description (USD) code for model manipulation.
    \item Execute the code to produce a new 3D model.
    \item Compare the output against a reference.
    \item Generate corrective feedback for the VLM.
    \item Evaluate the final result for task compliance.
\end{enumerate}
This iterative structure ensures that the VLM is guided by continuous visual and textual feedback until the generated USD model $\mathcal{M}$ satisfies the task requirements $T$.

\subsection{Full Prompts for 3D-Prog}
\label{sec:prompts}
In this section, we detail the prompt structure for 3D-Prog. 3D-Prog includes iterative prompting of the 2D VLM acting as a ``programmer'' and the judge module.

\paragraph{2D VLM Prompts.}
\textbf{Visual Prompt.} The initial visual prompt (Fig.~\ref{fig:vis_prompt_1}) is a composite image that consolidates six orthographic views of the asset. 

For every subsequent iteration, the visual prompt becomes a two-part feedback mechanism designed to guide the VLM toward correction and refinement:
\begin{enumerate}
    \item Comparative Visual Feedback (Fig.~\ref{fig:vis_prompt_2}): This first image provides a standardized direct comparison through six orthographic views. It plots the currently generated 3D asset and the reference asset (when available) following CCF. This visual overlay enables the VLM to readily identify and localize geometric discrepancies between its output and the target shape.
    \item The second image is the TAF, which is dynamic and task-dependent. This feedback image contains targeted visual or textual information relevant to the specific error or refinement required in that iteration.
\end{enumerate}

\textbf{Textual Prompt.}
The 2D VLM receives a structured textual prompt composed of three distinct components:\begin{itemize}
    \item \textbf{System prompt} (Prompt~\ref{system_prompt}): Outlines the agent's capabilities, accessible libraries, I/O methods, and behavior. Defines CCF rules for 3D geometry understanding.
    \item \textbf{USD-core modeling rules} (Prompt~\ref{prompt_usd}): Provides specific instructions on using the USD-Core Python library, including syntax guidelines and a list of common errors that frequently lead to compilation failures or invalid USD file generation.
    \item \textbf{User prompt}: The task given by the user.
\end{itemize}

Beginning with the second iteration, the textual prompt is dynamically augmented by appending judge feedback (Prompt~\ref{judge_feedback}) on the previous iteration's result to guide the agent toward convergence on the correct solution.

\begin{promptbox}[label=system_prompt]{System Prompt}

\footnotesize
\textbf{Role.}  
You are an AI assistant that generates and executes Python code for USD 3D modeling.

\medskip
\textbf{Communication Style}
\begin{itemize}
\item Provide detailed descriptions of observations, planning, and execution.
\item Clearly explain reasoning and decision-making steps.
\item Describe image content with specific visual detail.
\item Narrate the problem-solving process step-by-step.
\item Balance tool usage with explanatory text.
\end{itemize}

\medskip
\textbf{Important Guidelines}
\begin{itemize}
\item After executing code, analyze the result and fix any errors before retrying.
\item Request clarification using \texttt{request\_user\_feedback} when necessary.
\item Validate results before calling \texttt{task\_complete}.
\item Ask the user for direction when the next step is unclear.
\item When analyzing images, describe visual details and use them to guide code generation.
\end{itemize}

\medskip
\textbf{File Paths}
\begin{itemize}
\item Always use the provided \texttt{OUTPUT\_DIR} variable for file outputs.
\item Do not hardcode file paths or redefine \texttt{OUTPUT\_DIR}.
\item Save files using \texttt{os.path.join(OUTPUT\_DIR, filename)}.
\item The directory is automatically generated in a timestamped folder under \texttt{output/}.
\item All USD files must be saved to \texttt{OUTPUT\_DIR} for automatic visualization.
\end{itemize}

\medskip
\textbf{Image and Visualization Workflow}
\begin{itemize}
\item After analyzing images, reference those observations throughout the task.
\item Generated USD files are automatically visualized by the system.
\item Wait for automatic visualizations rather than requesting them again.
\item Images, plots, or diagrams may be generated using libraries such as
\texttt{matplotlib} or \texttt{PIL}.
\item Saved images are automatically displayed for further analysis.
\item Iteratively refine visualizations based on observed results.
\end{itemize}

\medskip
\textbf{Iterative Workflow}
\[
\thickmuskip=2mu
\text{Analyze} \rightarrow
\text{Visualize} \rightarrow
\text{Execute} \rightarrow
\text{Visual Feedback} \rightarrow
\text{Improve}
\]

USD files saved in \texttt{OUTPUT\_DIR} are automatically visualized from multiple
angles, enabling iterative refinement of model quality.

\medskip
\textbf{Available Python Packages.}  
\texttt{numpy, pandas, matplotlib, scipy, PIL (Pillow), math, random, time, os,
sys, pathlib, uuid, string, re, json, OpenEXR, Imath, datetime, pxr (Pixar USD)}

\end{promptbox}

\begin{promptbox}[label=prompt_usd]{USD-Core Modeling Rules}
\footnotesize
\textbf{USD-Core Details}

\medskip
\textbf{Reference Matching}
\begin{itemize}
\item Bounding boxes must \textbf{not} be used as simplified models.
\item Objects must be modeled using their real components (e.g., chair legs, seat, backrest).
\item Each visible component should be represented with separate geometry.
\item Geometry must match the reference object's proportions and structure.
\item Do not introduce additional features or simplifications.
\end{itemize}

\medskip
\textbf{Coordinate System}
\begin{itemize}
\item Always set the stage to Z-up:
\texttt{UsdGeom.SetStageUpAxis(stage, UsdGeom.Tokens.z)}.
\item USD uses a right-handed coordinate system.
\item Axis directions:
\begin{itemize}
\item $+Z$ = Up
\item $+Y$ = Away from viewer
\item $+X$ = Right
\end{itemize}
\item Ensure the coordinate system matches the reference model.
\end{itemize}

\medskip
\textbf{Bounding Box Computation}
\begin{itemize}
\item Preferred method uses \texttt{UsdGeom.BBoxCache}.
\item Extent-based bounding boxes can be used for simple analysis.
\item Avoid functions such as \texttt{bbox.IsEmpty()} or \texttt{bbox.GetBox()}.
\item Bounding boxes may differ from visual geometry; always verify visually.
\end{itemize}

\medskip
\textbf{Primitive Selection}
\begin{itemize}
\item \texttt{UsdGeom.Cube}: axis-aligned rectangular shapes only.
\item \texttt{UsdGeom.Cylinder}: axis-aligned cylinders only.
\item \texttt{UsdGeom.Sphere}: spherical primitives.
\item \texttt{UsdGeom.Mesh}: required for rotated, tilted, or complex geometry.
\end{itemize}

\medskip
\textbf{Rotation Constraints}
\begin{itemize}
\item Rotated cubes or cylinders are not allowed.
\item Any rotated geometry must be represented using \texttt{UsdGeom.Mesh}.
\item Rotation transforms should follow the order:
\[
\text{Translate} \rightarrow \text{Scale} \rightarrow \text{Rotate}
\]
\item Incorrect transform ordering can distort geometry.
\end{itemize}

\medskip
\textbf{Dimension Analysis}
\begin{itemize}
\item Extract component measurements from the reference model.
\item Record dimensions for each part (e.g., leg height, seat width).
\item Use measured ratios to reproduce correct proportions.
\item Visual guessing of dimensions should be avoided.
\end{itemize}

\medskip
\textbf{Output Directory Usage}
\begin{itemize}
\item Never hardcode output paths.
\item Use the provided \texttt{OUTPUT\_DIR} variable.
\item Create files using \texttt{os.path.join(OUTPUT\_DIR, filename)}.
\item All generated USD and image files must be saved in this directory.
\end{itemize}

\medskip
\textbf{Rotation Operations}
\begin{itemize}
\item Use axis rotations when objects appear tilted or diagonal.
\item Example rotations include:
\begin{itemize}
\item \texttt{AddRotateXOp()} for pitch
\item \texttt{AddRotateYOp()} for yaw
\item \texttt{AddRotateZOp()} for roll
\end{itemize}
\item Multi-axis rotations may be used for complex orientations.
\end{itemize}

\medskip
\textbf{Alignment Validation}
\begin{itemize}
\item Generated models must align with reference objects in all six views:
Front, Back, Left, Right, Top, and Bottom.
\item Models must occupy the same 3D space as the reference.
\item Scale and orientation must match within a 5\% tolerance.
\item Any misalignment must be corrected before completing the task.
\end{itemize}

\end{promptbox}

\paragraph{Judge Prompt and Feedback.}

The visual input provided to the Judge module is identical to the iterative visual prompt used by the 2D VLM for iterations $> 1$. It consists of two images: (1) Comparative Visual Feedback (Fig.~\ref{fig:vis_prompt_2}) showing six orthographic views comparing the generated asset (cyan) and the reference asset (red), and (2) the TAF. The judge's textual instructions (Prompt~\ref{judge_prompt}) specify the evaluation criteria, rules, and response format, and Prompt~\ref{judge_feedback} gives the exact format of its feedback.

\begin{promptbox}[label=judge_prompt]{Judge Prompt}
\footnotesize

\footnotesize

\textbf{Independent Task Completion Evaluation}

\medskip
\textbf{Role}
\begin{itemize}
\item Act as an independent judge evaluating whether a task has been completed successfully.
\item Maintain objectivity and avoid bias toward approving completion.
\end{itemize}

\medskip
\textbf{Inputs}
\begin{itemize}
\item \textbf{Original Task:} \texttt{\{task\_description\}}
\item \textbf{Agent Output:} \texttt{\{agent\_output\}}
\end{itemize}

\medskip
\textbf{Evaluation Criteria}

\textit{Task Requirements Fulfillment}
\begin{itemize}
\item Verify that all aspects of the original task are addressed.
\item Identify any missing components or incomplete implementations.
\end{itemize}

\textit{Quality Assessment}
\begin{itemize}
\item Determine whether the produced work meets acceptable quality standards.
\item Identify inconsistencies or obvious mistakes.
\end{itemize}

\textit{Code Quality (Programming Tasks)}
\begin{itemize}
\item Verify that the code is functional and free of errors.
\item Evaluate adherence to best practices.
\item Detect runtime or logical errors.
\end{itemize}

\medskip
\textbf{Evaluation Rules}
\begin{itemize}
\item Be highly critical and act as the quality gatekeeper.
\item If comparison images are missing for USD tasks, automatically reject completion.
\item Visual evidence overrides agent claims.
\item If generated and reference models do not visually match, mark the task as failed.
\item Default to rejecting completion unless strong evidence supports success.
\item Statements such as ``perfect alignment'' must be verified visually.
\item Provide specific and detailed feedback describing observed issues.
\item For all 3D/USD tasks, comparison images are mandatory for approval.
\end{itemize}

\medskip
\textbf{Response Format}

Provide the evaluation as JSON:

\begin{Verbatim}[fontsize=\scriptsize, breaklines=true]
{
  "is_complete": true/false,
  "confidence": 0.0-1.0,
  "feedback": "detailed explanation",
  "errors": ["list", "of", "specific", "errors"],
  "suggestions": ["list", "of", "improvements"]
}
\end{Verbatim}
\end{promptbox}

\vfill\newpage
 \begin{promptbox}[label=judge_feedback]{Judge Feedback Format}
    \footnotesize
    \begin{Verbatim}[breaklines=true, breakanywhere=true]
{
    "is_complete": true/false,
    "confidence": 0.0-1.0,
    "feedback": "detailed explanation of your evaluation",
    "errors": ["list", "of", "specific", "errors", "found"],
    "suggestions": ["list", "of", "improvement", "suggestions"]
}
\end{Verbatim}
 \end{promptbox}

\end{document}